\PassOptionsToPackage{table}{xcolor}
\documentclass{fairmeta}

\usepackage{amsmath,amsfonts,bm}

\def\eqref#1{equation~\ref{#1}}

\def\1{\bm{1}}

\DeclareMathAlphabet{\mathsfit}{\encodingdefault}{\sfdefault}{m}{sl}
\SetMathAlphabet{\mathsfit}{bold}{\encodingdefault}{\sfdefault}{bx}{n}

\usepackage{url}
\usepackage{amsmath,amssymb}
\usepackage{wrapfig}
\usepackage{tabularx}
\usepackage{makecell}
\usepackage{colortbl}
\usepackage{pifont}
\usepackage[ruled,linesnumbered]{algorithm2e}
\setcitestyle{round,authoryear}

\usepackage{adjustbox}

\headerleft{\includegraphics[height=15pt]{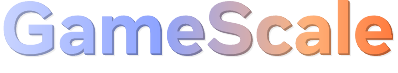}}
\headerright{\includegraphics[height=22pt]{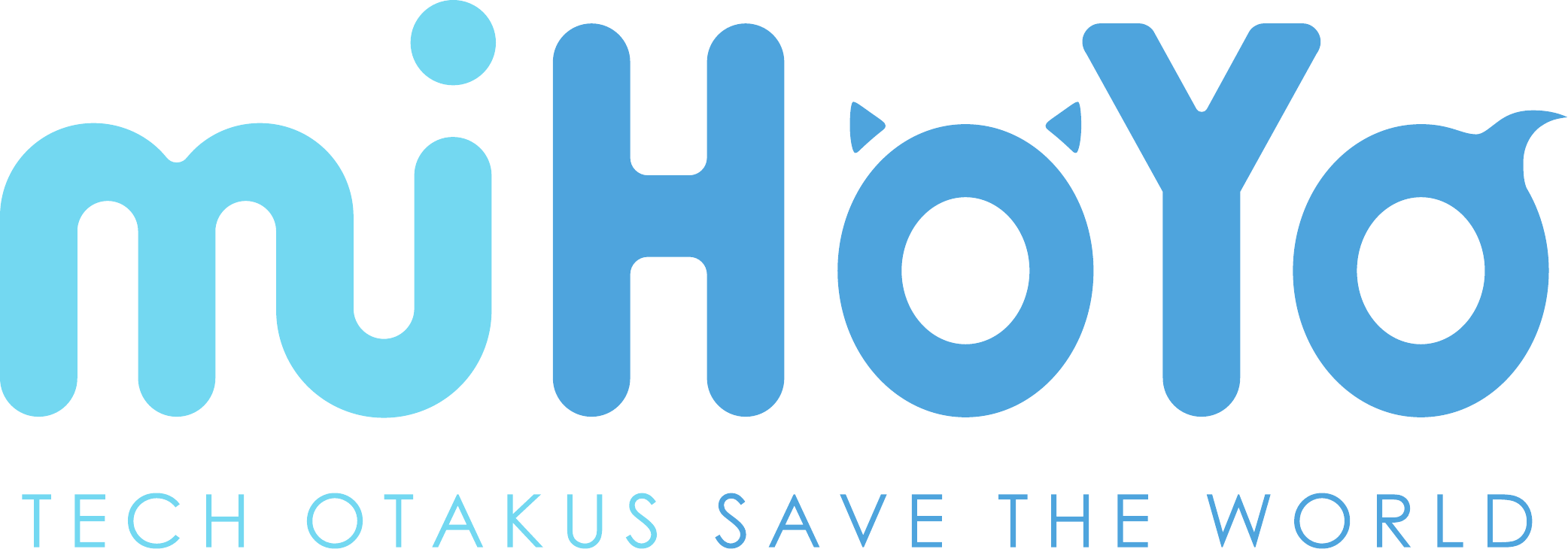}\hspace{8pt}\includegraphics[height=20pt]{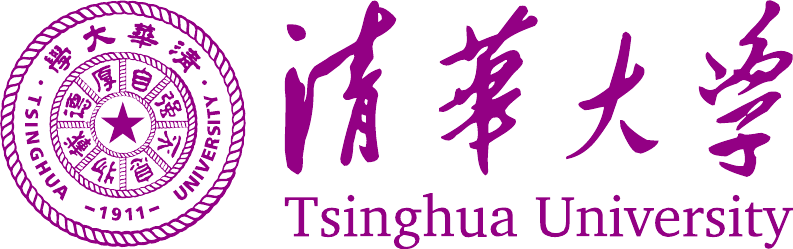}\hspace{8pt}\includegraphics[height=20pt]{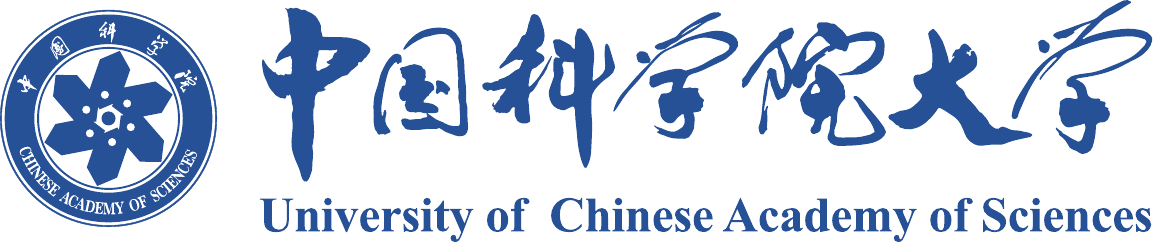}}

\title{\Large Scaling in Games: Continued Pre-Training For Embodied Agents In Diverse Virtual Worlds}

\author[1,2,*,\ddagger]{Kuan Zhang}
\author[1,3,*,\ddagger]{Yukun Chen}
\author[1,3,*,\ddagger]{Zhihao Yang}
\author[4]{Yue Su}
\author[3]{Xiangnan Wu}
\author[2]{Zirong Chen}
\author[1,5,\ddagger]{Run Luo}
\author[6]{Jinkun Hou}
\author[1]{Tao Tan}
\author[1,\dagger]{Yinhe Zheng}
\author[2,\dagger]{Yiming Li}
\renewcommand\affiliationformat[2][]{{\small $^{#1}$#2}}
\affiliation[1]{miHoYo Honkai AI R\&D Team}
\affiliation[2]{College AI, Tsinghua University}
\affiliation[3]{University of Chinese Academy of Sciences}
\affiliation[4]{MMLab, The University of Hong Kong}
\affiliation[5]{National University of Singapore}
\affiliation[6]{Peking University}
\contribution[*]{Equal Contribution}
\contribution[\dagger]{Corresponding Author}
\contribution[\ddagger]{Work done while interning at miHoYo}

\abstract{Scaling data, diversity and model capacity has improved embodied learning in physical environments, where diverse experiences remain grounded in shared physical laws. Whether these gains extend to virtual worlds with different visual styles, dynamics, controls, and objectives remains unclear. We investigate this question through a controlled study of continued pre-training (CPT) across three axes: data volume, diversity, and model size. Our 120 training configurations span four model sizes from 0.8B to 27B, six data budgets from 50 to 1,600 hours, and five mixtures ranging from Minecraft-only data to diverse multi-world corpora.  We assess performance and generalization using three-level validation losses and gameplay benchmarks from in-corpus worlds to unseen worlds. Our results show: (1) more data improves in-corpus performance, diversity improves transfer to related worlds with limited gains in distant worlds, and model size improves data efficiency but generalizes worse to distant worlds under single-world training; (2) we observe an early \emph{interface phase} with broad cross-world benefits and a later \emph{world phase} with gains narrowed to in-corpus and related worlds, and fit an empirical scaling law relating validation loss to data volume, diversity and model size; and (3) our models are competitive with Minecraft specialists and frontier models in 3D worlds, and perform best in real-time settings. However, the limits of imitation and persistent gaps in distant worlds suggest that building generalist embodied agents requires richer training stages beyond continued pre-training.
\vspace{-18pt}}

\metadata[Project Page]{\url{https://gamescaling.github.io}}
\metadata[Model]{\url{https://huggingface.co/THUSI-Lab/GameScaling}}
\metadata[Code]{\url{https://github.com/THUSI-Lab/GameScaling}}
\metadata[Correspondence]{\email{yimingli9702@mail.tsinghua.edu.cn}, \email{yinhe.zheng@mihoyo.com}}

\begin{document}

\maketitle

\section{Introduction}
\label{sec:intro}

Embodied agents are expected to perceive, reason, and act across changing environments. Recent scaling studies in robotics suggest that broader experience can improve such generality~\citep{ilscaling2025,rethinkingvla2026,crossembodiment2026}. Real-world scenes share one coherent physical reality, i.e., gravity, contact, and object permanence remain consistent, and the physical effect of an embodiment's controls changes relatively little across tasks. This shared structure makes experience collected in one setting potentially reusable in another.

\begin{figure*}[t]
\centering
\includegraphics[width=0.95\linewidth]{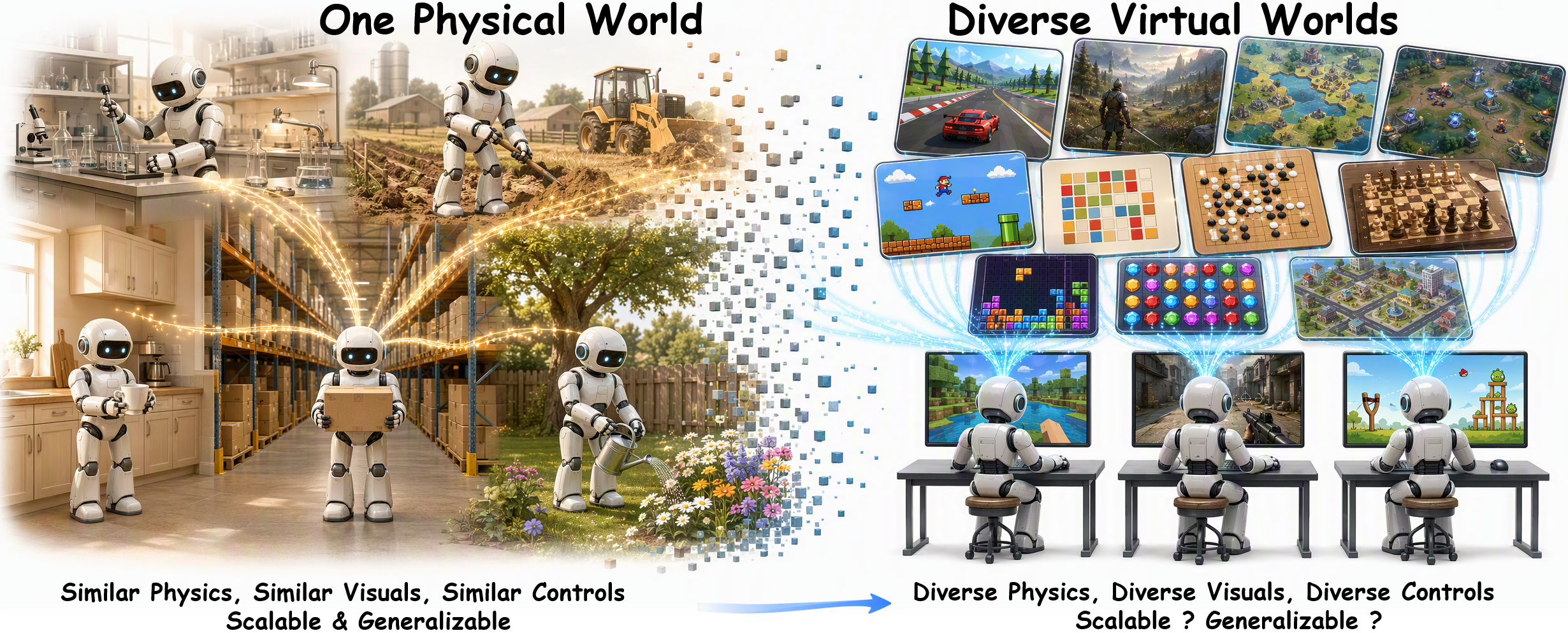}
\caption{\textbf{One physical world versus diverse virtual worlds.}
\emph{Left:} Agents learn from varied physical scenes sharing common dynamics, where they can scale and generalize with broader experience~\citep{rethinkingvla2026}.
\emph{Right:} Agents play diverse games with varying dynamics. We want to investigate the question that whether continued pre-training scales and generalizes in virtual worlds.}
\label{fig:teaser}
\end{figure*}

In Figure~\ref{fig:teaser}, virtual worlds remove much of this consistency. Different worlds may vary simultaneously in entirely different visuals, rules, physics, action spaces and goals~\citep{gametars2025, lumine2025, zhang2026multiverse}. Humans adapt to the heterogeneity quickly: they easily figure out unfamiliar rules, adjust to different controls, and apply past experience to new visuals and mechanics. A generalist virtual embodied agent should exhibit the same ability. Interactive video games provide a diverse and controlled subset of such worlds, making them a particularly demanding test of embodied generality~\citep{sima2024, sima2, zhang2026gameverse}.

Recent work studies this broader goal by training models to perceive visual input from games and output native keyboard--mouse controls across different virtual game worlds~\citep{baker2022video,nitrogen2026,yue2026scaling}. Across these systems, \textit{\textbf{continued pre-training}} (CPT) on gameplay trajectories has become a common stage for generalist embodied intelligence~\citep{sima2024,gametars2025,sima2,lumine2025}. However, virtual embodied continued pre-training remains largely a recipe: existing studies change the data volume, diversity and models together, leaving the effect of each factor unresolved. Most importantly, it remains unclear whether heterogeneous experiences help to build embodied agents that can generalize across different worlds, or if it just causes interference that ruins what they learned in any individual worlds.

We address this question with a controlled grid study spanning four model sizes from 0.8B to 27B, six data budgets from 50 to 1,600 hours, and five data mixture settings from Minecraft-only to broad multi-game corpora. The resulting \textbf{120 configurations} are evaluated through three-level validation loss and benchmark ladders from in-corpus worlds, such as Minecraft~\citep{zheng2025mcu} and Doom~\citep{Wydmuch2019ViZdoom}, to unseen near and distant worlds, such as OmniGameArena~\citep{omnigamearena2026}, and GameWorld~\citep{gameworld2026}. Our study reveals how data volume, mixture diversity, and model capacity affect embodied agents in these three levels of worlds, and whether heterogeneous experience becomes complementary as it scales.

Our contributions are mainly three-fold:
\begin{itemize}
   \item \textbf{A controlled study of scaling CPT in virtual worlds.}
    We quantify how data volume, mixture diversity, and model size affect CPT of generalist embodied agents in virtual worlds.
    More relevant data improves performance on in-corpus worlds where model performance multiplies in Minecraft. Diversity improves transfer to related worlds but provides limited gains in distant worlds. Larger models improve data efficiency and achieve larger early gains, while generalizing worse under single-game training. 

    \item \textbf{An empirical two-phase learning dynamics and scaling law.}
    We observe an empirical early \emph{interface phase}: model learns basic self-control and action primitives, which are shared and improved across worlds; and a \emph{world phase} then follows: further gains are obtained in trained and unseen related worlds.
    We fit an empirical scaling law to relate validation loss to data volume, mixture diversity and model size, achieving $R^2=0.982$ and $R^2=0.995$ on Minecraft and in-corpus loss, respectively.

    \item \textbf{Three-level evaluation of model performance and generalization.}
    We evaluate models in a three-level setting covering in-corpus, related, and distant worlds. This anchors our models among game and general foundation models, providing the community a reliable reference. Our models are competitive with Minecraft specialists and frontier models in 3D worlds, and outperform frontier models in real-time settings by up to $5\times$.

\end{itemize}

\section{Related Work}
\label{sec:related_work}

\textbf{Virtual Embodied Models.} Early models were trained as environment-specific specialists. Deep reinforcement learning achieved superhuman performance in Atari, Go, StarCraft II, and Dota 2 while relying on a fixed environment, objective, and action interface~\citep{mnih2015human,silver2016mastering,vinyals2019alphastar,berner2019dota}. Recent work instead focuses on large foundation models. VPT learns a Minecraft control prior from internet video, while NitroGen and OpenP2P extend pixel-to-action learning across broader game corpus~\citep{baker2022video,nitrogen2026, yue2026scaling}. LLM and VLM based agents further reuse pre-trained language and visual knowledge for instruction following, planning and control~\citep{sima2024, combatvla,gametars2025,lumine2025, sima2}. Across these approaches, CPT has become a common mechanism for turning a general backbone into an interactive model for virtual embodied agents. To our knowledge, we conduct the largest controlled CPT study of virtual embodied models.

\textbf{Virtual Worlds as Test Environments.} Games provide a controllable subset of virtual worlds for evaluating perception, reasoning, and action under diverse rules and dynamics~\citep{gameworld2026, zhang2026gameverse, GameHorizonSuite2026, zhang2026multiverse}. Minecraft and Doom suites measure capabilities within a familiar in-corpus world~\citep{Wydmuch2019ViZdoom, zheng2025mcu}; OmniGameArena tests transfer to unseen 3D worlds~\citep{omnigamearena2026}; and GameWorld extends evaluation to visually and mechanically distinct 2D, 2.5D, and 3D environments~\citep{gameworld2026}. Training in one kind of virtual worlds may transfer to others, but there is no previous study to investigate which factor is critical to the transfer.

\textbf{Scalability of Continued Pre-training.} Classical scaling laws relate model performance to capacity, data, and compute, while studies of domain-adaptive and repeated pre-training show that source composition can alter both adaptation and retention~\citep{kaplan2020scaling,hoffmann2022compute,dcpt2024,cmr2024,sparse_scaling2026}. Related evidence appears in computer-use and embodied learning: GUI studies find that task composition can matter as much as sample count~\citep{scalecua2026,guimid2025,datarecipes2026}, and robotic imitation studies report diversity scaling, negative transfer, and capacity-dependent interference~\citep{ilscaling2025,rethinkingvla2026,crossembodiment2026}. Virtual worlds sharpen this problem because visual style, dynamics, controls, and objectives change across sources. Existing research on CPT in virtual worlds alters model size, training data, and duration simultaneously. No prior work systematically isolates these three axes while measuring held-out loss and transfer capabilities at varying distances from training.

\section{Preliminaries and Experiment Framework}
\label{sec:setup}

\subsection{Data construction}
\label{sec:preliminaries}

\begin{figure}[t]
\centering
\includegraphics[width=\linewidth]{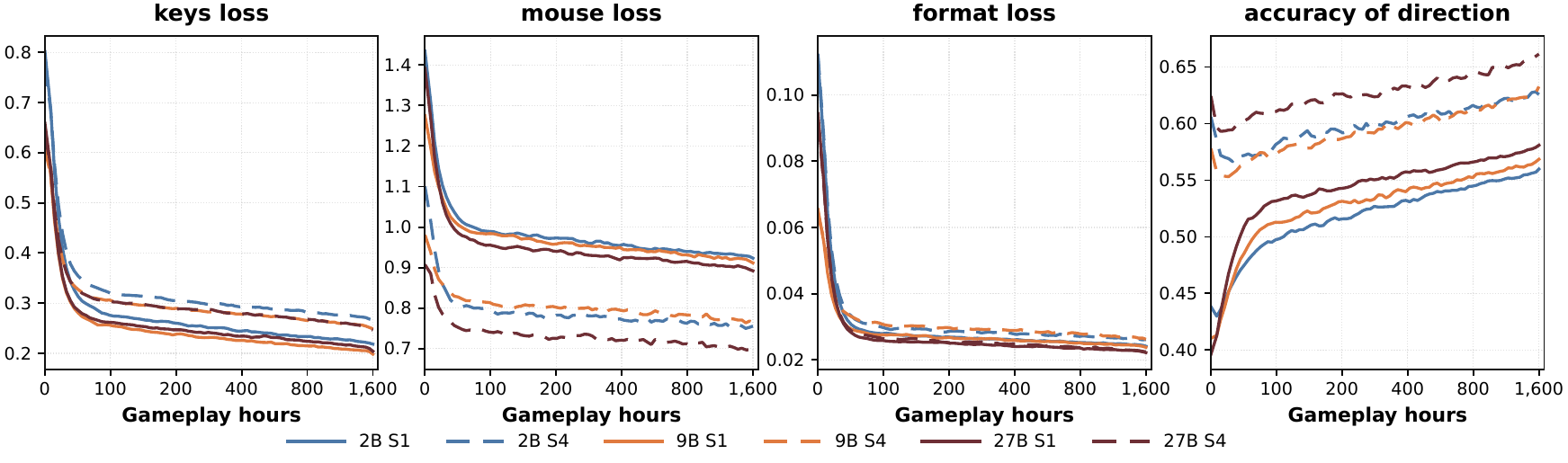}
\caption{\textbf{Adaptation to the keyboard--mouse interface during CPT.} Smoothed training losses for key, mouse, and format tokens, and eight-direction mouse prediction accuracy for three model capacities trained on mixtures S1 and S4.}

\label{fig:training_components}
\end{figure}

\textbf{Gameplay trajectories.}
Following the setting in previous works~\citep{gametars2025,lumine2025, sima2, omnigamearena2026}, we formulate gameplay as visual decision-making process. At step $t$, the model takes prompt context $c$, the current observation $o_t$, and interaction history $\mathcal{H}_t$ as the input $x_t$ and produces a reasoning trace $r_t$ and then an executable action $a_t$:
\begin{equation}
    r_t \sim p_\theta(\cdot\mid x_t),
    \qquad
    a_t \sim p_\theta(\cdot\mid x_t,r_t),
    \qquad
    x_t=(c,o_t,\mathcal{H}_t).
\end{equation}
Reasoning and action have distinct roles. The possibly empty $r_t$ is internal textual reasoning. The mandatory $a_t$ is structured text chunks that are parsed into low-level device commands, which further changes the environment. Each chunk covers 200\,ms~\citep{lumine2025, omnigamearena2026}:
\begin{equation}
    a_t =
    \texttt{<|action\_start|>}~\Delta x_t ~ \Delta y_t ~ \Delta z_t;
    K_t^{(1)};\ldots;K_t^{(6)}~\texttt{<|action\_end|>},
\end{equation}
where $(\Delta x_t,\Delta y_t)\in[-1000,1000]^2$ are discretized relative mouse moves, $\Delta z_t\in[-5,5]$ is the scroll delta, and each $K_t^{(j)}$ contains 0-4 keys or mouse buttons held for 33\,ms. The model receives observations at 5\,Hz and controls at 30\,Hz. History is modeled as interleaved image–action turns. Each trajectory has 20 turns. During training, preceding think and action are visible to later turns.

\textbf{Three scaling axes.}
Our study evaluates a $4\times6\times5=120$ configuration grid across four Qwen3.5 backbones $\mathcal{M}=\{\text{0.8B},\text{2B},\text{9B},\text{27B}\}$~\citep{qwen35}, six data budgets $\mathcal{T}=\{50,100,200,400,800,1600\}$, and five data-diversity mixtures $\mathcal{S}=\{\mathrm{S1},\ldots,\mathrm{S5}\}$. This yields total $3.5\times10^{11}$ training tokens and $2.2\times10^{22}$ FLOPs via the ``$6ND$ rule''~\citep{hoffmann2022compute}.

\textbf{Data diversity mixtures.}
Our training data consists of two components: (1) 85\% gameplay trajectories sampled from VPT, D2E, and OpenP2P~\citep{baker2022video,choi2025d2e,yue2026scaling}, and (2) 15\% general data covering vision--language, embodied and egocentric settings~\citep{li2024llava,li2026zebracot,tong2026gamerl,zhang2025embodiedreasoner,yang2025cambrians}. We construct five data mixtures ($\mathrm{S1}$--$\mathrm{S5}$) with increasing mixture diversity by setting the Minecraft trajectory proportion in the gameplay data to 100\%, 51.4\%, 27.1\%, 2.8\%, and 0\%, respectively. The remaining trajectories are sampled from a diverse pool of other games. We also augment a small fraction of gameplay turns with retrospective reasoning generated by GPT-5.6-Luna~\citep{gpt56card} based on the recorded observation and action, to preserve the backbone's native reasoning capabilities without altering action targets. See Appendices~\ref{app:dataset_construction} and~\ref{app:prompts} for further details about scaling on mixture diversity.

\subsection{Training and Evaluation}
\label{sec:training_evaluation}

\textbf{Training objective and validation sets.}
All models are continued pre-trained using a causal next-token prediction loss over assistant responses conditioned on interleaved histories. All parameters, including those of the ViT, are updated using AdamW~\citep{loshchilov2019decoupled} with a Warmup--Stable--Decay schedule~\citep{hu2024minicpm}. The resulting models are evaluated on three levels of validation data: Minecraft-only data, data from in-corpus games covered during training, and data from unseen games not covered in training. We also record key, mouse, and format losses. Apart from data volume, data diversity, and the choice of base model, all experimental settings are held fixed across training configurations. See Appendix~\ref{app:hyperparameters} for further details.

\textbf{Three-level gameplay evaluation.}
We evaluate gameplay competence across three categories of virtual worlds that are increasing distant from the training distribution. (1) \textbf{In-corpus worlds}: we use the MCU benchmark~\citep{zheng2025mcu} to evaluate performance on multiple tasks in Minecraft, and ViZDoom~\citep{Wydmuch2019ViZdoom} to evaluate performance on single shooting tasks in DOOM. Both games are covered in our training corpus. (2) \textbf{Unseen related worlds}: we evaluate our model in OmniGameArena (OGA)~\citep{omnigamearena2026}, which consists of seven unseen 3D worlds. Although their scenes and tasks are absent from training, action primitives such as navigation, aiming, jumping, shooting, and interaction have semantic counterparts in the training corpus. (3) \textbf{Unseen distant world}: we evaluate our model in GameWorld (GW)~\citep{gameworld2026}, which primarily covers 2D and 2.5D games, alongside a small portion of 3D games. These games differ substantially from the training data in visual styles, physics, controls and objectives. 

\section{Scaling and Generalization in Heterogeneous Virtual Worlds}
\label{sec:cpt_scaling}

\begin{figure*}[t]
\centering
\includegraphics[width=\textwidth]{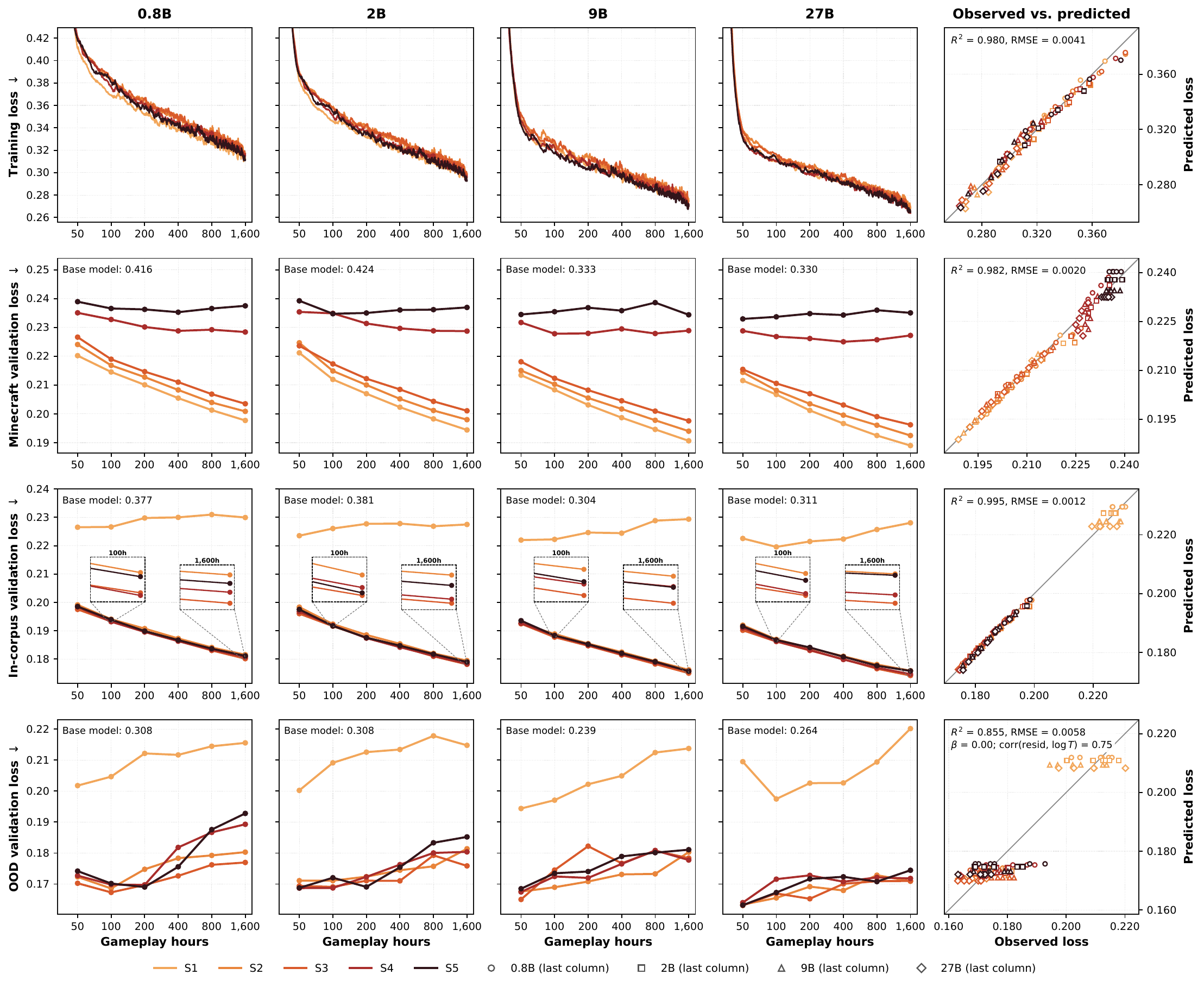}
\caption{\textbf{Loss scaling across data volume, model size and data diversity.} Rows report the smoothed training loss and three-level validation loss. Columns 1--4 denote the 0.8B-27B backbones; colors denote the training mixtures S1--S5; markers denote the six checkpoints from 50 to 1,600 hours. The base model loss is annotated in each validation panel. All plots in a given row shares a common y-axis scale. Column 5 plots the loss predicted by the fitted scaling law (Section~\ref{sec:scaling_law}) against the observed loss.}
\label{fig:loss_scaling_grid}
\end{figure*}

\begin{figure*}[t]
\centering
\includegraphics[width=\textwidth]{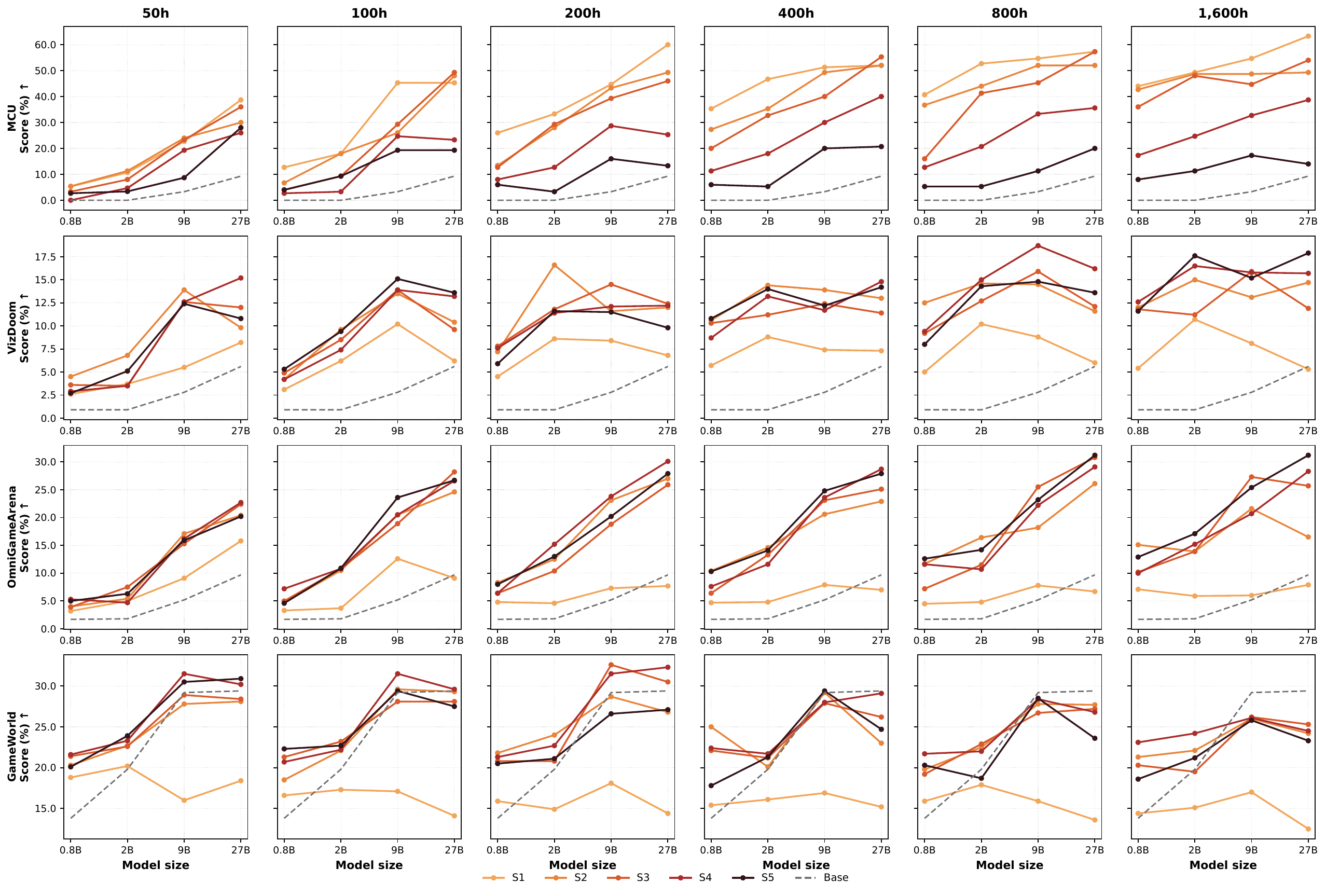}
\caption{\textbf{Benchmark performance across data budget, model size and data diversity.} Rows report MCU, ViZDoom, OmniGameArena, and GameWorld performance. Columns denote the six checkpoints from 50 to 1,600 hours. Colors denote S1-S5 and the base-model reference, while each curve spans the 0.8B-27B backbones. All plots in a given row shares a common y-axis scale.}
\label{fig:benchmark_grid}
\end{figure*}

We investigate how three scaling axes: data volume, data diversity, and model size, affect performance and generalization across all 120 configurations. Figure~\ref{fig:training_components} illustrates the action component losses. Figure~\ref{fig:loss_scaling_grid} presents the three-level training and validation losses alongside our scaling law predictions. Figure~\ref{fig:benchmark_grid} details benchmark evaluations. The next three subsections concentrate on each axis and the last subsection derives an empirical scaling law that unifies them.

\subsection{Data Scaling}
\label{sec:data_scaling}

\textbf{Early training brings broad cross-world gains.}
Figure~\ref{fig:training_components} shows the training loss of key, mouse, and format tokens. These losses drop rapidly during the initial tens of hours of training data before tapering into a more gradual decrease. Mouse direction prediction accuracy continues to improve. After exposure to 50 hours of training data, validation losses drop below those of the base models on all three validation sets (Figure \ref{fig:loss_scaling_grid}), even for games absent from the training mixture. We refer to this early stage as the \emph{interface phase}. This cross world improvement indicates that the model gradually acquires shared formats, representations and action patterns that can transfer across games.

\textbf{Later gains depend more on the worlds seen during training.}
After the interface phase, further performance gains depend heavily on exposure to relevant game worlds. We term this stage the \emph{world phase}. From 50 to 1,600 hours, the mean Minecraft validation loss decreases by about 10.4\% across four model sizes under the MineCraft-heavy mixtures S1--S3, correlating with substantial gains in MCU performance (Figure \ref{fig:loss_scaling_grid}). Conversely, scaling total data volume without sufficient Minecraft trajectories yields negligible further reduction in this specific validation loss. For the in-corpus validation set, diverse multi-game training consistently reduces loss across all model sizes, whereas Minecraft-only training causes the loss to plateau or even increase. Similarly, out-of-distribution (OOD) loss stagnates or rises, although it remains below  base model levels. At the largest model size, GameWorld performance degrades across all five mixtures as data volume increases. Thus, more data improves performance on in-corpus and related environments, but fails to drive continued gains in distant worlds.

\subsection{Diversity Scaling}
\label{sec:diversity_scaling}

\textbf{The best training mixture depends on the evaluation set.}
Across all 24 configurations of model size and data volume, Minecraft validation loss is lowest with S1, whereas in-corpus loss is minimized with S3 or S4 (Figure \ref{fig:loss_scaling_grid}). Increasing the Minecraft data proportion specifically improves Minecraft evaluation, whereas diverse mixtures improve general in-corpus performance. For in-corpus validation loss, differences among S2-S5 are negligible compared to their gap from S1. Remarkably, the zero-Minecraft S5 mixture still achieves a lower Minecraft validation loss than the base models. Comparisons at similar amounts of Minecraft data also show benefits from other games. 
Under a limited data budget, our results suggest matching the training distribution to the target evaluation domain rather than blindly expanding game diversity. Among all diversity mixtures, S1 has the highest in-corpus and OOD validation losses, along with the lowest ViZDoom, OGA, and GameWorld scores. Despite Minecraft encompassing diverse mechanics like combat and navigation, this within-game diversity fails to effectively generalize to other distinct virtual worlds.

\textbf{Diversity mainly improves transfer to related worlds.}
At 1,600 hours, the mean OGA score across S2--S5 is roughly 2.8 times that of S1, with all 16 mixed configurations outperforming S1 at matched model scales. ViZDoom performance similarly improves despite DOOM data forming only a small fraction of the training mixture. This indicates that basic navigation and aiming abilities can easily transfer across 3D games. A smaller model trained on diverse games can outperform a larger model trained mostly on Minecraft. On GameWorld, models trained with S1 degrade as training data volume increases. Although multi-game training mitigates this decline, the models still fail to consistently surpass their base models. Therefore, broader training diversity enhances capabilities in related environments but yields inconsistent gains in distant worlds.

\subsection{Model Scaling}
\label{sec:model_scaling}

\textbf{Larger models use data more efficiently and achieve larger early gains.} At limited data budgets, larger models generally show greater improvements on MCU and OGA. They achieve comparable or superior performance with up to a 32$\times$ less gameplay data. Large models trained on 50-hour data outperform small models trained on 1,600-hour data. These findings suggest that larger models extract more value from limited experience, requiring substantially less data to reach given performance level. However, further scaling yields uneven returns: Minecraft loss and ViZDoom performance exhibit diminishing gains beyond 9B, OGA does not improve uniformly across all mixtures, and CPT adds little to the existing GameWorld performance of larger backbones.

\textbf{Single-game training can amplify cross-world decline in larger models.}
During the world phase, model size interacts with the training mixture. Under mixed-game training, OOD loss becomes more stable as models grow; Under single-game training (S1), the late-stage degradation in OOD performance is most severe for the largest model. In-corpus loss reflects a similar weakening of the larger models' advantage under S1. At 1,600 hours, 27B-S1 underperforms its base model on all three non-Minecraft benchmarks, suffering a maximum performance drop on GameWorld of roughly 17 points. Thus, greater data efficiency under single-game training can actively compromise cross-world generalization. Broadening the training data mitigates this loss, though generalization to distant worlds remains a challenge.

\subsection{An Empirical Three-Factor Scaling Law on CPT Loss}
\label{sec:scaling_law}

Following D-CPT~\citep{dcpt2024}, we formulate an empirical scaling law relating validation loss to data volume, diversity and model size. Specifically, the Chinchilla scaling law $L=E+A/N^\alpha+B/T^\beta$~\citep{hoffmann2022compute} is extended by introducing a ratio $\rho$ into the data term to account for gameplay diversity. Moreover, we observe that even without Minecraft data, the S5 mixture yields a finite Minecraft validation loss well below that of the base model. To capture this cross-mixture transfer, we add a term that remains finite even at a zero data share:
\begin{equation}
L(N,T,\rho)
=E+\frac{A}{N^{\alpha}}
+\frac{B\rho^{\eta}}{T^{\beta}}
+\frac{C}{(\rho+\epsilon)^{\gamma}},
\label{eq:three_factor}
\end{equation}
where $N$ is the parameter count, $T$ is gameplay hours, and $E$, $A$, $\alpha$, $B$, $\eta$, $\beta$, $C$, $\gamma$, $\epsilon>0$ are fitted parameters. Let $r$ denote the proportion of Minecraft trajectories in the training mixture. We set $\rho=r$ when predicting Minecraft validation loss, and $\rho=1-r$ for in-corpus loss. Finally, because each mixture is evaluated on its respective training distribution, the training loss itself is fitted using the standard form $L=E_S+A/N^\alpha+B/T^\beta$.

The empirical scaling laws yield strong fits for training, Minecraft, and in-corpus losses, achieving $R^2$ values of $0.980$, $0.982$, and $0.995$, with RMSE of $0.0041$, $0.0020$, and $0.0012$, respectively. The fits also extrapolate: refitted without the 27B models, the 1,600-hour checkpoints or the 50-hour checkpoints, Eq.~(\ref{eq:three_factor}) predicts the held-out validation losses with RMSE between $0.0016$ and $0.0027$. OOD loss, in contrast, does not follow a scaling law. Its gain over the base model is already reached at 50 hours, after which it rises slightly with further training in all 20 curves. Fitting Eq.~(\ref{eq:three_factor}) to it therefore gives $\beta\approx0$, and the remaining $R^2$ of $0.855$ only reflects static differences between mixtures. Coupled with the marginal CPT gains on GameWorld in Figure~\ref{fig:benchmark_grid}, this suggests that CPT improves learning within the corpus but does not keep improving on distant worlds. Fitting details, the derivation of Eq.~(\ref{eq:three_factor}), comparisons with alternative models and uncertainty analyses are given in Appendix~\ref{app:scaling_law_experiments}.

\section{Comparing with Frontier Models}
\label{sec:benchmark_comparison}

\begin{table*}[t]
    \centering
    \caption{\textbf{Comparison with baselines on MCU.} We report average steps and success rate (ASR, \%) across 30 comprehensive tasks. Baselines are quoted from GameWAM. Within each model group, the \textcolor{red}{red} and \textcolor{blue}{blue} indicate the best and second, respectively. ``--'' denotes an unavailable result.}
    \label{tab:mcu_model_comparison}
    \renewcommand\arraystretch{1.15}
    \providecommand{\best}[1]{\textcolor{red}{#1}}
    \providecommand{\second}[1]{\textcolor{blue}{#1}}
    \adjustbox{max width=\textwidth}{%
    \begin{tabular}{@{}lc cc cc cc cc@{}}
    \toprule
    \multirow{2}{*}{Model} & \multirow{2}{*}{Size} &
    \multicolumn{2}{c}{\includegraphics[height=1.2em]{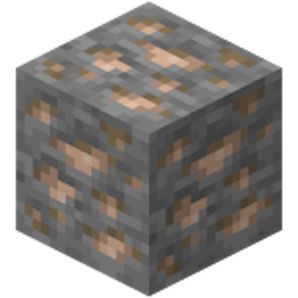} \textbf{Embodied Tasks}} &
    \multicolumn{2}{c}{\includegraphics[height=1.2em]{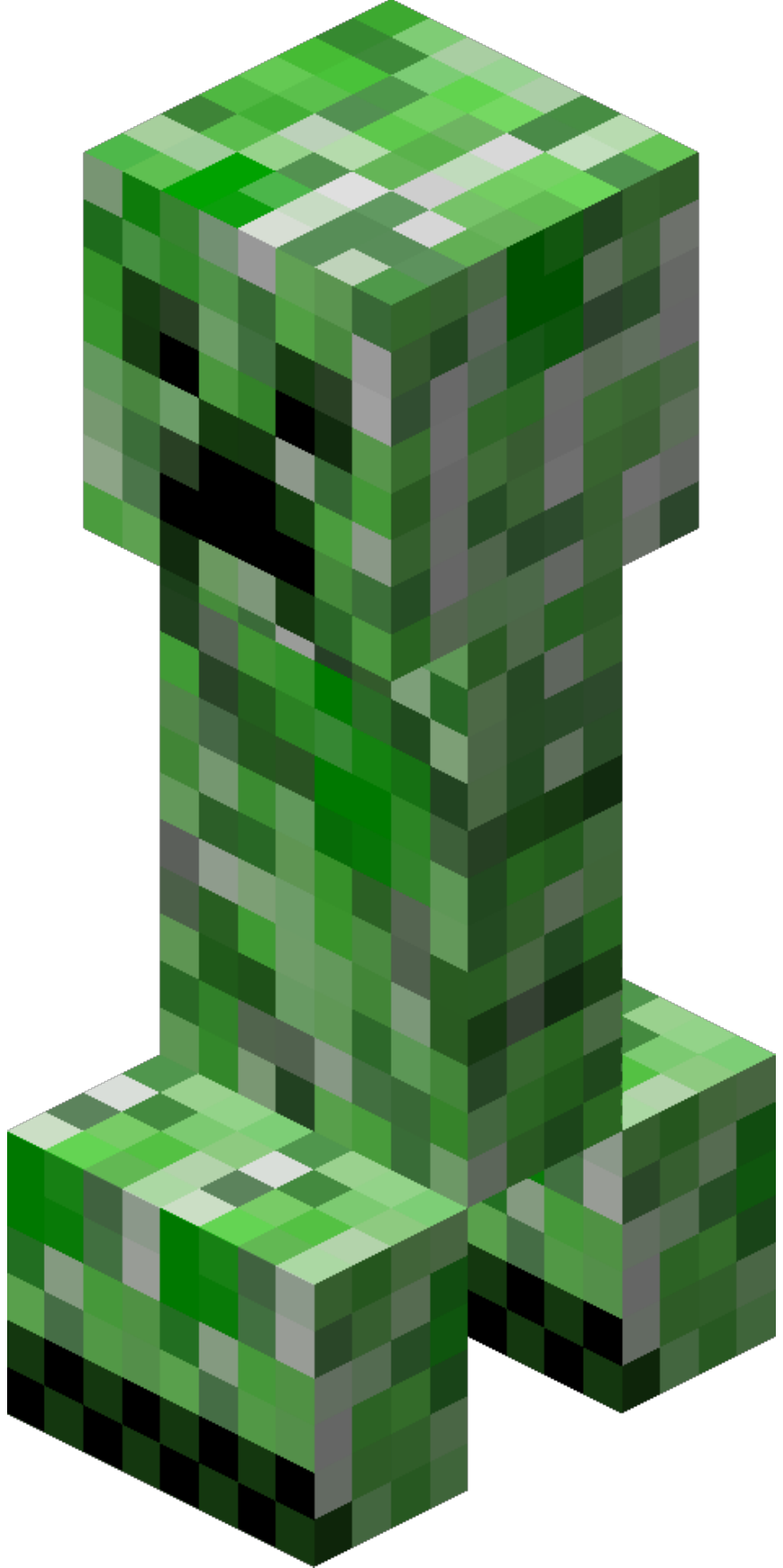} \textbf{Combat Tasks}} &
    \multicolumn{2}{c}{\includegraphics[height=1.2em]{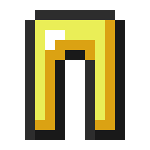} \textbf{GUI Tasks}} &
    \multicolumn{2}{c}{\textbf{Average}} \\
    \cmidrule(lr){3-4}\cmidrule(lr){5-6}\cmidrule(lr){7-8}\cmidrule(l){9-10}
    & & Steps $\downarrow$ & ASR $\uparrow$
    & Steps $\downarrow$ & ASR $\uparrow$
    & Steps $\downarrow$ & ASR $\uparrow$
    & Steps $\downarrow$ & ASR $\uparrow$ \\
    \midrule
    \rowcolor{gray!15}
    \multicolumn{10}{l}{\textbf{\textit{Environment-Specific Model: Trained on MineCraft trajectories}}} \\
    VPT~\citep{baker2022video} & 248M & 377 & $10.1^{\pm3.6}$ & 396 & $3.6^{\pm7.7}$ & 398 & $0.7^{\pm0.1}$ & 390 & 4.8 \\
     STEVE-1~\citep{lifshitz2023steve} & 248M & 384 & $8.4^{\pm3.0}$ & 395 & $4.9^{\pm1.8}$ & 391 & $0.0$ & 390 & 4.4 \\
    ROCKET-1~\citep{cai2025rocket} & 72B & 392 & $19.2^{\pm6.1}$ & 320 & $29.8^{\pm9.0}$ & -- & -- & 356 & 24.5 \\
    JARVIS-VLA~\citep{li2025jarvis} & 7B & 305 & $31.0^{\pm12.7}$ & 352 & $18.3^{\pm5.2}$ & 339 & $25.3^{\pm5.7}$ & 332 & 24.9 \\
    OpenHA~\citep{wang2025openha} & 7B & 287 & $37.0^{\pm15.9}$ & 316 & $40.0^{\pm19.6}$ & 314 & $33.3^{\pm13.3}$ & 306 & 36.8 \\
    MAIN-VLA~\citep{zhou2026mainvla} & 7B & 263 & $38.5^{\pm11.8}$ & \second{248} & $49.3^{\pm13.5}$ & \second{291} & $36.7^{\pm8.1}$ & \second{267} & 41.5 \\
    \textbf{GameScaling-0.8B-S1} & 0.8B & \second{255} & $74.0^{\pm34.7}$ & 374 & $50.0^{\pm33.8}$ & 590 & $8.0^{\pm13.3}$ & 406 & 44.0 \\
    \textbf{GameScaling-2B-S1} & 2B & 263 & \best{$80.0^{\pm23.7}$} & 335 & $52.0^{\pm32.5}$ & 553 & $16.0^{\pm12.0}$ & 384 & 49.3 \\
    GameWAM~\citep{gamewam} & 6B & \best{138} & $70.0^{\pm32.2}$ & \best{203} & $39.0^{\pm25.9}$ & \best{155} & \second{$43.0^{\pm32.9}$} & \best{165} & 50.7 \\
    \textbf{GameScaling-9B-S1} & 9B & 319 & $66.0^{\pm28.4}$ & 307 & \second{$58.0^{\pm37.4}$} & 483 & $40.0^{\pm35.8}$ & 370 & \second{54.7} \\
    \textbf{GameScaling-27B-S1} & 27B & 272 & \second{$76.0^{\pm21.5}$} & 320 & \best{$64.0^{\pm33.2}$} & 422 & \best{$50.0^{\pm39.2}$} & 338 & \best{63.3} \\
    \midrule
    \rowcolor{gray!15}
    \multicolumn{10}{l}{\textit{\textbf{General Game Foundation Model: Trained across diverse games}}} \\
    NitroGen~\citep{nitrogen2026} & 500M & 581 & $0.5^{\pm0.1}$ & 510 & $2.1^{\pm0.3}$ & 594 & $0.3^{\pm0.0}$ & 562 & 1.0 \\
    Open-P2P~\citep{yue2026scaling} & 1.2B & 489 & $1.5^{\pm0.6}$ & 466 & $5.3^{\pm1.5}$ & 586 & $0.4^{\pm0.1}$ & 514 & 2.4 \\
    \textbf{GameScaling-0.8B-S3} & 0.8B & 345 & $68.0^{\pm31.2}$ & 420 & $38.0^{\pm31.6}$ & 596 & $2.0^{\pm6.0}$ & 454 & 36.0 \\
    GameTARS-Dense~\citep{gametars2025} & 7B & 373 & $50.4^{\pm20.7}$ & 372 & $38.1^{\pm24.6}$ & \second{406} & $39.1^{\pm27.5}$ & 384 & 42.5 \\
    \textbf{GameScaling-9B-S3} & 9B & 329 & $66.0^{\pm28.4}$ & 425 & $36.0^{\pm28.0}$ & 512 & $32.0^{\pm29.9}$ & 422 & 44.7 \\
    \textbf{GameScaling-2B-S3} & 2B & \second{260} & \best{$76.0^{\pm19.6}$} & 355 & $52.0^{\pm25.6}$ & 563 & $16.0^{\pm12.0}$ & 393 & 48.0 \\
    \textbf{GameScaling-27B-S3} & 27B & 326 & $62.0^{\pm22.7}$ & \second{339} & \second{$58.0^{\pm30.3}$} & 460 & \second{$42.0^{\pm22.7}$} & \second{375} & \second{54.0} \\
    GameTARS-MoE~\citep{gametars2025} & 20B-active MoE & \best{188} & \second{$72.0^{\pm21.5}$} & \best{271} & \best{$66.1^{\pm25.2}$} & \best{288} & \best{$55.4^{\pm17.6}$} & \best{249} & \best{64.5} \\
    \bottomrule
    \end{tabular}%
    }
\end{table*}
This section compares our experimentally derived models (labeled as GameScaling in Tables \ref{tab:mcu_model_comparison} and \ref{tab:cross_benchmark_reference}) against current frontier models. By doing so, we establish a comprehensive performance reference across four distinct virtual benchmarks.

\textbf{Comparison with Minecraft game models.}
Table~\ref{tab:mcu_model_comparison} shows the performance of our model on MCU~\citep{zheng2025mcu}. Existing baselines fall into two categories: environment-specific models built exclusively for Minecraft, from controllers trained on videos~\citep{baker2022video,lifshitz2023steve,cai2025rocket} to instruction-tuned backbones~\citep{li2025jarvis,wang2025openha,zhou2026mainvla}; and general game models trained across diverse games environments using a unified keyboard--mouse interface, either with~\citep{gametars2025} or without~\citep{nitrogen2026,yue2026scaling} a language backbone. Based on training data, our S1 models align with the former group and S3 models with the latter; critically, neither relies on task-specific supervision.

We highlight three key findings. First, generative backbone capability outweighs Minecraft-specific design: our 27B-S1 model outperforms the best specialized model by 13 points, and even the 0.8B-S1 surpasses 7B specialists. Second, a language foundation is more critical than sheer data scale: behavior-cloning models trained on tens of thousands of hours score near zero on instruction-conditioned tasks, while our 0.8B-S3 completes 36\% of tasks. Third, task categories exhibit distinct scaling behaviors. Success rates on embodied tasks remain stable across our model sizes and lead all baselines, whereas GUI task success increases sixfold from 0.8B to 27B. This suggests that while coarse movement is acquired early in the interface phase; precise cursor control necessitates substantial model capacity. 

\begin{table*}[t]
\centering
\caption{\textbf{Performance across four benchmarks under two interaction settings.} For proprietary models whose size label follows the approximate estimate~\citep{modelsizefitting}. TP and RT denote the time-paused and real-time settings. The best is marked in \textcolor{red}{red} and the second in \textcolor{blue}{blue}.}
\label{tab:cross_benchmark_reference}
\adjustbox{max width=\textwidth}{%
\renewcommand\arraystretch{1.15}%
\begin{tabular}{@{}lc cc cc cc cc cc@{}}
\toprule
\multirow{2}{*}{Model} & \multirow{2}{*}{Size} &
\multicolumn{2}{c}{MCU} &
\multicolumn{2}{c}{ViZDoom} &
\multicolumn{2}{c}{OmniGameArena} &
\multicolumn{2}{c}{GameWorld} &
\multicolumn{2}{c}{Average} \\
\cmidrule(lr){3-4}\cmidrule(lr){5-6}\cmidrule(lr){7-8}\cmidrule(lr){9-10}\cmidrule(l){11-12}
& & TP & RT
& TP & RT
& TP & RT
& TP & RT
& TP & RT \\
\midrule
\rowcolor{gray!15}
\multicolumn{12}{l}{\textit{\textbf{Proprietary Frontier Models}}} \\
Gemini-3.8-Flash & $\sim$T & \textcolor{red}{88.7} & 7.3 & \textcolor{red}{27.8} & 2.8 & \textcolor{blue}{45.2} & \textcolor{red}{18.6} & \textcolor{red}{60.7} & \textcolor{red}{48.2} & \textcolor{red}{55.6} & \textcolor{red}{19.2} \\
GPT-5.6-Sol & $\sim$T & \textcolor{blue}{82.0} & \textcolor{red}{8.7} & \textcolor{blue}{26.7} & 2.7 & \textcolor{red}{45.6} & 14.5 & \textcolor{blue}{53.9} & \textcolor{blue}{44.3} & \textcolor{blue}{52.1} & \textcolor{blue}{17.6} \\
GPT-5.6-Luna & $\sim$T & 54.0 & 5.3 & 24.1 & \textcolor{red}{2.9} & 26.7 & 13.2 & 44.9 & 41.6 & 37.4 & 15.8 \\
Claude-Opus-5.0 & $\sim$T & 56.7 & \textcolor{red}{8.7} & 20.8 & 2.0 & 44.4 & \textcolor{blue}{16.4} & 52.8 & 42.0 & 43.7 & 17.3 \\
\rowcolor{gray!15}
\multicolumn{12}{l}{\textit{\textbf{Open-Source Frontier Models}}} \\
GLM-5.3-Flash & 320B & 47.3 & 7.3 & 18.9 & 1.9 & 23.5 & 12.3 & 34.0 & 29.6 & 30.9 & 12.8 \\
DeepSeek-V4-Flash-Vision & 305B & 36.0 & 7.3 & 13.0 & 3.3 & 27.1 & 11.6 & 41.8 & 28.5 & 29.5 & 12.7 \\

\midrule
\rowcolor{gray!15}
\multicolumn{12}{l}{\textit{\textbf{Open-Source Qwen Models and GameScaling Models}}} \\
Qwen3.5-0.8B & 0.8B & 0.0 & 0.0 & 0.9 & 1.0 & 1.7 & 1.8 & 13.8 & 15.1 & 4.1 & 4.5 \\
\textbf{GameScaling-0.8B-S3} & 0.8B & 36.0 & 37.3 & 11.9 & 11.8 & 10.2 & 9.8 & 20.3 & 22.5 & 19.6 & 20.4 \\
Qwen3.5-2B & 2B & 0.0 & 0.0 & 0.9 & 1.5 & 1.8 & 1.7 & 19.8 & 16.3 & 5.6 & 4.9 \\
\textbf{GameScaling-2B-S3} & 2B & \textcolor{blue}{48.0} & \textcolor{red}{44.0} & 11.2 & 10.4 & 13.9 & 12.2 & 19.5 & 20.0 & 23.2 & \textcolor{blue}{21.7} \\
Qwen3.5-9B & 9B & 3.3 & 3.3 & 2.8 & 3.2 & 5.2 & 4.1 & \textcolor{blue}{29.2} & \textcolor{blue}{27.7} & 10.1 & 9.6 \\
\textbf{GameScaling-9B-S3} & 9B & 44.7 & \textcolor{blue}{36.7} & \textcolor{red}{15.9} & \textcolor{red}{16.4} & \textcolor{red}{27.3} & \textcolor{red}{20.2} & 26.2 & \textcolor{red}{27.8} & \textcolor{blue}{28.5} & \textcolor{red}{25.3} \\
Qwen3.5-27B & 27B & 9.3 & 4.0 & 5.6 & 2.9 & 9.7 & 7.6 & \textcolor{red}{29.3} & 25.2 & 13.5 & 9.9 \\
\textbf{GameScaling-27B-S3} & 27B & \textcolor{red}{54.0} & 16.7 & \textcolor{blue}{14.7} & \textcolor{blue}{13.2} & \textcolor{blue}{25.7} & \textcolor{blue}{17.8} & 25.3 & 24.8 & \textcolor{red}{29.9} & 18.1 \\
\bottomrule
\end{tabular}}
\end{table*}

\textbf{Cross-benchmark reference.}
Table~\ref{tab:cross_benchmark_reference} compares proprietary frontier models~\citep{gpt56card,google2026gemini38flash,anthropic2026opus5}, open-source frontier models~\citep{qwen35,glm53flash,deepseekv4flashvision}, and our S3 models using a unified keyboard--mouse interface. We evaluate under two settings: 1. time-paused interaction, which isolates pure reasoning competence, and 2. real-time interaction, which assesses competence and inference latency simultaneously. Under time-paused interaction, performance strongly depends on generalization distance. In 3D worlds, CPT amplifies the 27B backbone's score by roughly $6\times$ on MCU and $2.6\times$ on OGA; notably, our 9B and 27B models match the performance of $\sim$300B open models. 

In the distant GameWorld benchmark, frontier models score twice as high as ours, and CPT only yields improvements at 0.8B and 2B scales, highlighting distant-world transfer as the primary bottleneck for generality. Under real-time interaction, this performance hierarchy reverses. On ViZDoom and MCU, every frontier model suffers an approximate 90\% score reduction because delayed decisions are executed in an environment state that has already evolved. Our 0.8B-9B models retain 88\% of their time-paused performance, outscoring the best frontier models several times over. On OGA, which features exploration-heavy tasks, this real-time penalty for frontier models is roughly halved. Ultimately, the real-time penalty scales with the tempo of the world, and a model two orders of magnitude smaller is the better player when the world does not pause.

\section{Discussion and Conclusion}
\label{sec:discussion}

\subsection{Scalability and Generalization Depend on Virtual Worlds}

The observed two-phase learning dynamics clarify why scaling reliably enhances in-domain gameplay but struggles to achieve broad transfer. Early training yields cross-worlds improvements, reflecting the acquisition of universal controls and action patterns. Subsequent gains, however, depend more on the training distribution. By 1,600 hours, diverse mixed training produces a mean OGA score about $2.8\times$ higher than Minecraft-only training, averaged across model sizes (Section~\ref{sec:diversity_scaling}). These benefits fail to generalize consistently to distant virtual worlds. Even a feature-rich environment like Minecraft, which encompasses combat, exploration, and navigation, yields limited transfer when used in isolation. Consequently, scaling model capacity or training duration must be paired with strategic curation of the training corpus. With a limited data budget, the training mixture should closely mirror the target evaluation environments. Finally, while empirical scaling laws accurately estimate in-corpus losses, a strong in-domain fit does not guarantee robust play in distant worlds.

\subsection{Games Require Both Ability and Speed}

Frontier models have advanced on gaming benchmarks (Figure~\ref{fig:frontier_trends}), and our GameScaling models prove highly competitive with both Minecraft specialists and frontier models in 3D environments (Section~\ref{sec:benchmark_comparison}). Crucially, their relative performance hinges heavily on whether the environment pauses during inference. Under real-time evaluation, GameScaling-9B-S3 achieves an average score of 25.3 across the four benchmarks, outperforming the best-tested frontier model (19.2) (Table~\ref{tab:cross_benchmark_reference}). In dynamic unpaused worlds, a delayed action often becomes obsolete by the time it is executed. Real-time competence therefore depends on both gameplay capability and response latency. Further, these metrics reflect real-world deployment realities: GameScaling uses vLLM API endpoints on our H200 cluster, whereas frontier models incur network latency via provider APIs. Future comparisons must transparently report task scores, response times, and deployment infrastructure to accurately assess a model's true utility during active gameplay.

\subsection{Toward a General Game Model}

While CPT establishes a strong foundation for generalist game agents, performance gains in related environments do not necessarily translate to distant worlds. For instance, compared to its base model, the 27B-S3 checkpoint improves the time-paused OGA score by 16.0 points but suffers a 4.0-point drop on GameWorld (Table~\ref{tab:cross_benchmark_reference}). This discrepancy underscores the necessity of distant-world evaluation to accurately assess model generality. CPT relies on behavior cloning from recorded actions, lacking direct environmental feedback regarding task success or failure. The observed generalization gap may stem from this imitation-based objective, insufficient game coverage, or a combination of both. Future work should investigate whether scaling world diversity, integrating online feedback during actual play, or combining both paradigms enhances cross-world transfer~\citep{zhang2026gameverse}.

\section*{Ethics Statement}

Our gameplay training data are sampled from the publicly released VPT, D2E, and OpenP2P datasets~\citep{baker2022video,choi2025d2e,yue2026scaling} under their respective licenses. These datasets were cleaned by their providers to protect player privacy. We sample existing gameplay trajectories and add retrospective reasoning annotations, without collecting new gameplay recordings or personal data. We identify no additional ethical concerns arising from these procedures.

\clearpage
\bibliographystyle{unsrtnat}
\bibliography{main}

@article{zhang2026gameverse,
  title={{GameVerse}: Can Vision-Language Models Learn from Video-based Reflection?},
  author={Zhang, Kuan and Liu, Dongchen and Zhao, Qiyue and Hou, Jinkun and Zhang, Xinran and Xie, Qinlei and others},
  journal={arXiv preprint arXiv:2603.06656},
  year={2026}
}

@article{zhang2026multiverse,
  title={Towards Generalist Game Players: An Investigation of Foundation Models in the Game Multiverse},
  author={Zhang, Kuan and Liu, Dongchen and Zhao, Qiyue and Xin, Tianyu and Su, Yue and Wang, Haisheng and others},
  journal={arXiv preprint arXiv:2605.09965},
  year={2026}
}

@article{mnih2015human,
  title={Human-level control through deep reinforcement learning},
  author={Mnih, Volodymyr and Kavukcuoglu, Koray and Silver, David and Rusu, Andrei A. and Veness, Joel and Bellemare, Marc G. and others},
  journal={Nature},
  volume={518},
  number={7540},
  pages={529--533},
  year={2015},
  doi={10.1038/nature14236}
}

@article{Wydmuch2019ViZdoom,
  author  = {Marek Wydmuch and Micha{\l} Kempka and Wojciech Ja\'skowski},
  title   = {{ViZDoom} {C}ompetitions: {P}laying {D}oom from {P}ixels},
  journal = {IEEE Transactions on Games},
  year    = {2019},
  volume  = {11},
  number  = {3},
  pages   = {248--259},
  doi     = {10.1109/TG.2018.2877047},
  note    = {The 2022 IEEE Transactions on Games Outstanding Paper Award}
}

@article{silver2016mastering,
  title={Mastering the game of {Go} with deep neural networks and tree search},
  author={Silver, David and Huang, Aja and Maddison, Chris J. and Guez, Arthur and Sifre, Laurent and van den Driessche, George and others},
  journal={Nature},
  volume={529},
  number={7587},
  pages={484--489},
  year={2016},
  doi={10.1038/nature16961}
}

@article{vinyals2019alphastar,
  title={Grandmaster level in {StarCraft II} using multi-agent reinforcement learning},
  author={Vinyals, Oriol and Babuschkin, Igor and Czarnecki, Wojciech M. and Mathieu, Micha{\"e}l and Dudzik, Andrew and Chung, Junyoung and others},
  journal={Nature},
  volume={575},
  number={7782},
  pages={350--354},
  year={2019},
  doi={10.1038/s41586-019-1724-z}
}

@article{berner2019dota,
  title={Dota 2 with Large Scale Deep Reinforcement Learning},
  author={{OpenAI} and Berner, Christopher and Brockman, Greg and Chan, Brooke and Cheung, Vicki and D{\k{e}}biak, Przemys{\l}aw and others},
  journal={arXiv preprint arXiv:1912.06680},
  year={2019}
}

@inproceedings{baker2022video,
  title={Video PreTraining ({VPT}): Learning to Act by Watching Unlabeled Online Videos},
  author={Baker, Bowen and Akkaya, Ilge and Zhokhov, Peter and Huizinga, Joost and Tang, Jie and Ecoffet, Adrien and others},
  booktitle={Advances in Neural Information Processing Systems},
  volume={35},
  pages={24639--24654},
  year={2022}
}

@inproceedings{lifshitz2023steve,
  title={{STEVE-1}: A Generative Model for Text-to-Behavior in Minecraft},
  author={Lifshitz, Shalev and Paster, Keiran and Chan, Harris and Ba, Jimmy and McIlraith, Sheila},
  booktitle={Advances in Neural Information Processing Systems},
  volume={36},
  year={2023}
}

@inproceedings{cai2025rocket,
  title={{ROCKET-1}: Mastering Open-World Interaction with Visual-Temporal Context Prompting},
  author={Cai, Shaofei and Wang, Zihao and Lian, Kewei and Mu, Zhancun and Ma, Xiaojian and Liu, Anji and others},
  booktitle={Proceedings of the IEEE/CVF Conference on Computer Vision and Pattern Recognition},
  pages={12122--12131},
  year={2025},
  doi={10.1109/CVPR52734.2025.01132}
}

@inproceedings{li2025jarvis,
  title={{JARVIS-VLA}: Post-Training Large-Scale Vision Language Models to Play Visual Games with Keyboards and Mouse},
  author={Li, Muyao and Wang, Zihao and He, Kaichen and Ma, Xiaojian and Liang, Yitao},
  booktitle={Findings of the Association for Computational Linguistics: ACL 2025},
  pages={17878--17899},
  publisher={Association for Computational Linguistics},
  year={2025},
  doi={10.18653/v1/2025.findings-acl.920}
}

@article{wang2025openha,
  title={{OpenHA}: A Series of Open-Source Hierarchical Agentic Models in Minecraft},
  author={Wang, Zihao and Li, Muyao and He, Kaichen and Wang, Xiangyu and Mu, Zhancun and Liu, Anji and others},
  journal={arXiv preprint arXiv:2509.13347},
  year={2025}
}

@article{zhou2026mainvla,
  title={{MAIN-VLA}: Modeling Abstraction of Intention and eNvironment for Vision-Language-Action Models},
  author={Zhou, Zheyuan and Du, Liang and Sun, Zixun and Zhou, Xiaoyu and Ye, Ruimin and Chen, Qihao and others},
  journal={arXiv preprint arXiv:2602.02212},
  year={2026}
}

@article{choi2025d2e,
  title={{D2E}: Scaling Vision-Action Pretraining on Desktop Data for Transfer to Embodied AI},
  author={Choi, Suhwan and Jung, Jaeyoon and Seong, Haebin and Kim, Minchan and Kim, Minyeong and Cho, Yongjun and others},
  journal={arXiv preprint arXiv:2510.05684},
  year={2025}
}

@article{gametars2025,
  title={{Game-TARS}: Pretrained Foundation Models for Scalable Generalist Multimodal Game Agents},
  author={Wang, Zihao and Li, Xujing and Ye, Yining and Fang, Junjie and Wang, Haoming and Liu, Longxiang and others},
  journal={arXiv preprint arXiv:2510.23691},
  year={2025}
}

@inproceedings{nitrogen2026,
  title={{NitroGen}: An Open Foundation Model for Generalist Gaming Agents},
  author={Magne, Lo{\"i}c and Awadalla, Anas and Wang, Guanzhi and Xu, Yinzhen and Belofsky, Joshua and Hu, Fengyuan and others},
  booktitle={Proceedings of the IEEE/CVF Conference on Computer Vision and Pattern Recognition},
  pages={21511--21521},
  year={2026}
}

@article{lumine2025,
  title={Lumine: An Open Recipe for Building Generalist Agents in 3D Open Worlds},
  author={Tan, Weihao and Li, Xiangyang and Fang, Yunhao and Yao, Heyuan and Yan, Shi and Luo, Hao and others},
  journal={arXiv preprint arXiv:2511.08892},
  year={2025}
}

@article{sima2024,
  title={Scaling Instructable Agents Across Many Simulated Worlds},
  author={Abi Raad, Maria and Ahuja, Arun and Barros, Catarina and Besse, Frederic and Bolt, Andrew and Bolton, Adrian and others},
  journal={arXiv preprint arXiv:2404.10179},
  year={2024}
}

@article{sima2,
  title={{SIMA 2}: A Generalist Embodied Agent for Virtual Worlds},
  author={{SIMA Team}},
  journal={arXiv preprint arXiv:2512.04797},
  year={2025}
}

@article{yue2026scaling,
  title={Scaling Behavior Cloning Improves Causal Reasoning: An Open Model for Real-Time Video Game Playing},
  author={Yue, Yuguang and Salia, Irakli and Hunt, Samuel and Green, Chris and Shi, Wenzhe and Hunt, Jonathan J.},
  journal={arXiv preprint arXiv:2601.04575},
  year={2026}
}

@article{gamewam,
  title={{GameWAM}: A World Action Model for Video Games},
  author={Guo, Yuncheng and Zhang, Zhanqiu and Guo, Yiwen and Li, Weijia},
  journal={arXiv preprint arXiv:2608.26200},
  year={2026}
}

@misc{google2026gemini38flash,
  title={{Gemini 3.8 Flash} Model Card},
  author={{Google DeepMind}},
  howpublished={Model card},
  year={2026},
  url={https://deepmind.google/models/model-cards/gemini-3-8-flash}
}

@misc{anthropic2026opus5,
  title={System Card: {Claude Opus 5}},
  author={{Anthropic}},
  howpublished={System card},
  year={2026},
  url={https://www.anthropic.com/claude-opus-5-system-card}
}

@article{li2024llava,
  title={{LLaVA-OneVision}: Easy Visual Task Transfer},
  author={Li, Bo and Zhang, Yuanhan and Guo, Dong and Zhang, Renrui and Li, Feng and Zhang, Hao and others},
  journal={Transactions on Machine Learning Research},
  year={2025}
}

@inproceedings{li2026zebracot,
  title={{Zebra-CoT}: A Dataset for Interleaved Vision-Language Reasoning},
  author={Li, Ang and Wang, Charles L. and Fu, Deqing and Yue, Kaiyu and Cai, Zikui and Zhu, Wang Bill and others},
  booktitle={International Conference on Learning Representations},
  year={2026}
}

@inproceedings{tong2026gamerl,
  title={{Game-RL}: Synthesizing Multimodal Verifiable Game Data to Boost {VLMs}' General Reasoning},
  author={Tong, Jingqi and Tang, Jixin and Li, Hangcheng and Mou, Yurong and Zhang, Ming and Zhao, Jun and others},
  booktitle={International Conference on Learning Representations},
  year={2026}
}

@inproceedings{zhang2025embodiedreasoner,
  title={{Embodied-Reasoner}: Synergizing Visual Search, Reasoning, and Action for Embodied Interactive Tasks},
  author={Zhang, Wenqi and Wang, Mengna and Liu, Gangao and Xu, Huixin and Jiang, Yiwei and Shen, Yongliang and others},
  booktitle={Proceedings of the 64th Annual Meeting of the Association for Computational Linguistics (Volume 1: Long Papers)},
  pages={41178--41207},
  publisher={Association for Computational Linguistics},
  year={2026}
}

@inproceedings{yang2025cambrians,
  title={{Cambrian-S}: Towards Spatial Supersensing in Video},
  author={Yang, Shusheng and Yang, Jihan and Huang, Pinzhi and Brown, Ellis and Yang, Zihao and Yu, Yue and others},
  booktitle={International Conference on Learning Representations},
  year={2026}
}

@inproceedings{zheng2025mcu,
  title={{MCU}: An Evaluation Framework for Open-Ended Game Agents},
  author={Zheng, Xinyue and Lin, Haowei and He, Kaichen and Wang, Zihao and Fu, Qiang and Fu, Haobo and others},
  booktitle={Proceedings of the 42nd International Conference on Machine Learning},
  series={Proceedings of Machine Learning Research},
  volume={267},
  pages={78221--78259},
  publisher={PMLR},
  year={2025}
}

@article{gameworld2026,
  title={{GameWorld}: Towards Standardized and Verifiable Evaluation of Multimodal Game Agents},
  author={Ouyang, Mingyu and Hu, Siyuan and Lin, Kevin Qinghong and Ng, Hwee Tou and Shou, Mike Zheng},
  journal={arXiv preprint arXiv:2604.07429},
  year={2026}
}

@article{omnigamearena2026,
  title={{OmniGameArena}: A Unified {UE5} Benchmark for {VLM} Game Agents with Improvement Dynamics},
  author={Lin, Mingxian and Qian, Shengju and Liu, Yuqi and Huang, Yi-Hua and Wang, Yiyu and Huang, Wei and others},
  journal={arXiv preprint arXiv:2606.09826},
  year={2026}
}

@inproceedings{loshchilov2019decoupled,
  title={Decoupled Weight Decay Regularization},
  author={Loshchilov, Ilya and Hutter, Frank},
  booktitle={International Conference on Learning Representations},
  year={2019}
}

@article{hu2024minicpm,
  title={{MiniCPM}: Unveiling the Potential of Small Language Models with Scalable Training Strategies},
  author={Hu, Shengding and Tu, Yuge and Han, Xu and He, Chaoqun and Cui, Ganqu and Long, Xiang and others},
  journal={arXiv preprint arXiv:2404.06395},
  year={2024}
}

@article{kaplan2020scaling,
  title={Scaling Laws for Neural Language Models},
  author={Kaplan, Jared and McCandlish, Sam and Henighan, Tom and Brown, Tom B. and Chess, Benjamin and Child, Rewon and others},
  journal={arXiv preprint arXiv:2001.08361},
  year={2020}
}

@inproceedings{hoffmann2022compute,
  title={An Empirical Analysis of Compute-Optimal Large Language Model Training},
  author={Hoffmann, Jordan and Borgeaud, Sebastian and Mensch, Arthur and Buchatskaya, Elena and Cai, Trevor and Rutherford, Eliza and others},
  booktitle={Advances in Neural Information Processing Systems},
  volume={35},
  pages={30016--30030},
  year={2022}
}

@inproceedings{scalecua2026,
  title={{ScaleCUA}: Scaling Open-Source Computer Use Agents with Cross-Platform Data},
  author={Liu, Zhaoyang and Xie, JingJing and Ding, Zichen and Li, Zehao and Yang, Bowen and Wu, Zhenyu and others},
  booktitle={International Conference on Learning Representations},
  year={2026}
}

@article{guimid2025,
  title={Breaking the Data Barrier -- Building {GUI} Agents Through Task Generalization},
  author={Zhang, Junlei and Ding, Zichen and Ma, Chang and Chen, Zijie and Sun, Qiushi and Lan, Zhenzhong and others},
  journal={arXiv preprint arXiv:2504.10127},
  year={2025}
}

@article{datarecipes2026,
  title={{OpenThoughts-Agent}: Data Recipes for Agentic Models},
  author={Raoof, Negin and Zhuang, Richard and Nezhurina, Marianna and Guha, Etash and Tejaswi, Atula and Marten, Ryan and others},
  journal={arXiv preprint arXiv:2606.24855},
  year={2026}
}

@inproceedings{ilscaling2025,
  title={Data Scaling Laws in Imitation Learning for Robotic Manipulation},
  author={Lin, Fanqi and Hu, Yingdong and Sheng, Pingyue and Wen, Chuan and You, Jiacheng and Gao, Yang},
  booktitle={International Conference on Learning Representations},
  year={2025}
}

@article{rethinkingvla2026,
  title={Rethinking Visual-Language-Action Model Scaling: Alignment, Mixture, and Regularization},
  author={Wang, Ye and Zheng, Sipeng and Luo, Hao and Zhang, Wanpeng and Yuan, Haoqi and Xu, Chaoyi and others},
  journal={arXiv preprint arXiv:2602.09722},
  year={2026}
}

@inproceedings{crossembodiment2026,
  title={Cross-Embodiment Offline Reinforcement Learning for Heterogeneous Robot Datasets},
  author={Abe, Haruki and Osa, Takayuki and Mukuta, Yusuke and Harada, Tatsuya},
  booktitle={International Conference on Learning Representations},
  year={2026}
}

@inproceedings{dcpt2024,
  title={{D-CPT Law}: Domain-specific Continual Pre-Training Scaling Law for Large Language Models},
  author={Que, Haoran and Liu, Jiaheng and Zhang, Ge and Zhang, Chenchen and Qu, Xingwei and Ma, Yinghao and others},
  booktitle={Advances in Neural Information Processing Systems},
  volume={37},
  pages={90318--90354},
  year={2024},
  doi={10.52202/079017-2867}
}

@inproceedings{cmr2024,
  title={{CMR} Scaling Law: Predicting Critical Mixture Ratios for Continual Pre-training of Language Models},
  author={Gu, Jiawei and Yang, Zacc and Ding, Chuanghao and Zhao, Rui and Tan, Fei},
  booktitle={Proceedings of the 2024 Conference on Empirical Methods in Natural Language Processing},
  pages={16143--16162},
  publisher={Association for Computational Linguistics},
  year={2024},
  doi={10.18653/v1/2024.emnlp-main.903}
}

@inproceedings{sparse_scaling2026,
  title={When Data Is Scarce: Scaling Sparse Language Models with Repeated Training},
  author={Wu, Boqian and Xiao, Qiao and Okanovic, Patrik and Sternal, Tomasz and van Keulen, Maurice and Pechenizkiy, Mykola and others},
  booktitle={Proceedings of the 43rd International Conference on Machine Learning},
  year={2026}
}

@misc{qwen35,
  title  = {{Qwen3.5}: Towards Native Multimodal Agents},
  author = {{Qwen Team}},
  month  = feb,
  year   = {2026},
  url    = {https://qwen.ai/blog?id=qwen3.5}
}

@misc{gpt56card,
  author = {{OpenAI}},
  title  = {{GPT-5.6} Preview System Card},
  year   = {2026},
  month  = jun,
  url    = {https://deploymentsafety.openai.com/gpt-5-6-preview},
  note   = {Official system card, published 26 June 2026; accessed 2026-07-31}
}

@misc{combatvla,
      title={CombatVLA: An Efficient Vision-Language-Action Model for Combat Tasks in 3D Action Role-Playing Games}, 
      author={Peng Chen and Pi Bu and Yingyao Wang and Xinyi Wang and Ziming Wang and Jie Guo and Yingxiu Zhao and Qi Zhu and Jun Song and Siran Yang and Jiamang Wang and Bo Zheng},
      year={2025},
      eprint={2503.09527},
      archivePrefix={arXiv},
      primaryClass={cs.CV},
      url={https://arxiv.org/abs/2503.09527}, 
}

@misc{glm53flash,
  title  = {{GLM-5.3-Flash}: Frontier Intelligence, Flash Cost},
  author = {{Z.ai}},
  year   = {2026},
  url    = {https://z.ai/blog/glm-5.3-flash},
  note   = {Technical blog and model card; the model follows the GLM-5 technical report}
}

@misc{deepseekv4flashvision,
  title  = {{DeepSeek-V4-Flash-Vision-Exp}},
  author = {{DeepSeek-AI}},
  year   = {2026},
  url    = {https://huggingface.co/deepseek-ai/DeepSeek-V4-Flash-Vision-Exp},
  note   = {Model card; builds on the DeepSeek-V4-Flash architecture with visual modules}
}

@article{modelsizefitting,
  title={Incompressible knowledge probes: Estimating black-box llm parameter counts via factual capacity},
  author={Li, Bojie},
  journal={arXiv preprint arXiv:2604.24827},
  year={2026}
}

@article{hill1973diversity,
  title={Diversity and Evenness: A Unifying Notation and Its Consequences},
  author={Hill, Mark O.},
  journal={Ecology},
  volume={54},
  number={2},
  pages={427--432},
  year={1973}
}

@article{jost2006entropy,
  title={Entropy and Diversity},
  author={Jost, Lou},
  journal={Oikos},
  volume={113},
  number={2},
  pages={363--375},
  year={2006}
}

@misc{markov2026gaming500,
  title  = {{Gaming-500-Hours}: Native Gameplay Screen Recordings with Input Events},
  author = {{Markov AI}},
  year   = {2026},
  howpublished = {Hugging Face dataset},
  url    = {https://huggingface.co/datasets/markov-ai/gaming-500-hours},
  note   = {494.7 hours of gameplay across 168 games}
}

@article{GameHorizonSuite2026,
      title={GameHorizon Suite: Multi-Horizon Data and Evaluation in Gameplay},
      author={Yiran Wang and Xingyilang Yin and Junfu Pu and Guangzhi Wang and Kaifeng Li and Mingyu Ouyang and Huiqiang Sun and Lingen Li and Cheng Cheng and Wangbo Yu and Honghao Chen and Xiaodong Cun and Chi-Man Pun and Zhiguo Cao and Ying Shan},
      year={2026},
      journal={arXiv preprint arXiv:2609.25001},
      url={https://arxiv.org/abs/2609.25001},
}

\clearpage
\beginappendix
\section{Training and Dataset Details}
\label{app:training}

\subsection{Dataset Construction and Mixture Accounting}
\label{app:dataset_construction}

All configurations allocate 85\% of training tokens to gameplay trajectories and 15\% to a fixed vision--language anchor. The gameplay pool combines VPT, D2E, and OpenP2P~\citep{baker2022video,choi2025d2e,yue2026scaling}. The vision--language anchor comprises LLaVA-OneVision, Zebra-CoT, GameQA, Embodied-Reasoner, and Cambrian-S data~\citep{li2024llava,li2026zebracot,tong2026gamerl,zhang2025embodiedreasoner,yang2025cambrians}. The general vision--language data does not include game content from the 85\% gameplay component. This prevents the fixed general vision–language data from providing additional exposure to these game worlds when we vary the diversity of the gameplay component. Its composition and token share remain unchanged across S1--S5. We vary data diversity by changing the balance between Minecraft and a fixed pool of other game worlds. Across S2--S5, the non-Minecraft worlds remain the same, and their relative proportions within that pool are approximately unchanged. This shifts training from Minecraft-dominated data toward a more balanced mix of 79 worlds, without progressively adding new non-Minecraft games. Holding the pool fixed helps separate the effect of world proportions from the effect of the number of worlds.  Running the 120-condition grid with one training seed required approximately 45,000 H200 GPU-hours. Given this cost, we leave repeated training with different seeds and experiments that vary the number of worlds to future work.

We normalize source-specific mouse coordinates, sample observations at 5\,Hz, and construct 20-turn windows with a 10-turn stride. Each window supervises 4.0 seconds of actions and spans 3.8 seconds between its first and last observations. We use the latter interval to measure gameplay duration. Actions are converted to the unified six-substep format, and idle-dominated windows are downsampled. For the heterogeneous pool, we sample complete recordings and cap each source stream to limit dominance by long streams. S2--S4 use nested subsets of this capped pool, while S5 follows the same procedure after removing Minecraft. So the measured Minecraft shares among gameplay segments are 100\%, 51.4\%, 27.1\%, and 2.8\%, respectively. S5 removes these residual Minecraft trajectories, reducing the share to 0\%.

\textbf{Fixed training-game inventory.}
S2--S5 draw from the same inventory of 124 non-Minecraft source streams; the mixtures change only how many segments are drawn from those streams. S2--S4 additionally contain Minecraft, whereas S5 removes both D2E and OpenP2P Minecraft streams. S1 contains only VPT Minecraft. The 124 shared streams correspond to 79 canonical non-Minecraft labels; adding Minecraft gives the 80 labels listed in Table~\ref{tab:training_game_inventory}. Thus, ``124'' refers to the fixed source-stream inventory, not to 124 title-disjoint commercial games. Two very small residual metadata labels (\emph{Roblox: Debug} and \emph{Other / Unlabeled}) are retained in the accounting table rather than silently reassigned.

\textbf{Mixture diversity.}
\label{app:mixture_diversity}
Because the world inventory is fixed, S1--S5 cannot be ordered by the number of worlds: S2, S3, and S4 contain the same 80 canonical labels. What changes is how gameplay experience is distributed over them. Let $p_w$ be the share of gameplay segments from canonical world $w$ in a mixture. We measure mixture diversity by the effective number of worlds,
\begin{equation}
D=\exp\Big(-\sum_w p_w\log p_w\Big),
\end{equation}
the Hill number of order one~\citep{hill1973diversity,jost2006entropy}. $D$ equals the number of equally weighted worlds that would give the same Shannon entropy. It combines how many worlds are present with how evenly experience is spread across them, so it is 1 for a single world and 80 for 80 equally weighted worlds. Table~\ref{tab:mixture_diversity} reports $D$ together with the inverse Simpson index $1/\sum_w p_w^2$, which weights dominant worlds more heavily, and the largest single-world share. Both indices order S1--S4 identically. Only the step from S1 to S2 enlarges the world set; from S2 to S4 the set is unchanged, and diversity grows because Minecraft's dominant share is reassigned to the heterogeneous worlds. S5 has nearly the same diversity as S4 and differs from it mainly by excluding the residual 2.8\% of Minecraft, so it serves as a source-free control rather than a further diversity step. At a fixed budget, raising $D$ also raises the hours of each heterogeneous world and lowers the Minecraft hours, because S2--S4 draw nested subsets from the same capped pool.

\begin{table}[!htbp]
\centering
\small
\caption{\textbf{Mixture diversity of S1--S5.} Shares are computed over gameplay segments after cross-source aliases are merged into canonical world labels. $D$ is the effective number of worlds (Hill number of order one); the inverse Simpson index emphasizes dominant worlds.}
\label{tab:mixture_diversity}
\begin{tabular}{lrrrrr}
\toprule
Mixture & Worlds & Minecraft share & Largest share & $D=\exp H$ & Inverse Simpson \\
\midrule
S1 & 1 & 100\% & 100\% & 1.0 & 1.0 \\
S2 & 80 & 51.4\% & 51.4\% & 14.5 & 3.7 \\
S3 & 80 & 27.1\% & 27.1\% & 35.2 & 11.9 \\
S4 & 80 & 2.8\% & 4.2\% & 60.1 & 52.6 \\
S5 & 79 & 0\% & 4.3\% & 58.7 & 51.0 \\
\bottomrule
\end{tabular}
\end{table}

We use the six game categories shown in the dataset table: AVG (adventure/survival), STG (shooting), RPG (role-playing), RAC (racing), SIM (simulation), and OTH (platform, sports, MOBA, and other games). The category assignment remain fixed across S1--S5.

\begin{table}[t]
\centering
\caption{\textbf{Fixed game inventory represented across S1--S5.} The table merges source aliases into canonical labels. $n$ is the number of distinct D2E/OpenP2P source-stream labels represented by that canonical label; the non-Minecraft entries sum to 124. VPT Minecraft is accounted for separately.}
\label{tab:training_game_inventory}
\scriptsize
\setlength{\tabcolsep}{3.5pt}
\adjustbox{max width=\linewidth}{%
\begin{tabular}{llr@{\hspace{12pt}}llr}
\toprule
\textbf{Canonical game} & \textbf{Category} & \textbf{$n$} &
\textbf{Canonical game} & \textbf{Category} & \textbf{$n$} \\
\midrule
Barony & AVG & 1 & GTA: San Andreas & RPG & 1 \\
Core Keeper & AVG & 1 & MapleStory Worlds Southperry & RPG & 1 \\
Dinkum & AVG & 1 & Monster Hunter Wilds & RPG & 1 \\
Grounded & AVG & 1 & Saints Row 4 & RPG & 1 \\
Minecraft & AVG & 2 & Saints Row: The Third & RPG & 1 \\
Ogu Forest & AVG & 1 & Skul & RPG & 1 \\
PEAK & AVG & 1 & Elefant Hovercraft & RAC & 5 \\
Raft & AVG & 1 & Need for Speed & RAC & 1 \\
Roblox: A Dusty Trip & AVG & 4 & Need for Speed: Carbon & RAC & 1 \\
Roblox: Natural Disaster Survival & AVG & 5 & Need for Speed: Most Wanted & RAC & 1 \\
Sea of Thieves & AVG & 1 & Need for Speed: Underground & RAC & 1 \\
Spider-Man 2 & AVG & 1 & Need for Speed: Underground 2 & RAC & 3 \\
Apex Legends & STG & 1 & DRL Simulator & SIM & 2 \\
Battlefield 6 Open Beta & STG & 1 & Euro Truck Simulator 2 & SIM & 4 \\
Brotato & STG & 1 & FPV Kamikaze Drone & SIM & 3 \\
Call of Duty: Black Ops Zombies & STG & 1 & House Flipper & SIM & 1 \\
Call of Duty Mobile & STG & 1 & House Flipper 2 & SIM & 1 \\
Call of Duty: Black Ops 2 & STG & 4 & Liftoff & SIM & 2 \\
Call of Duty: Black Ops 3 & STG & 2 & Medieval Dynasty & SIM & 1 \\
Counter-Strike 2 & STG & 1 & PowerWash Simulator & SIM & 2 \\
DOOM & STG & 7 & Satisfactory & SIM & 1 \\
DOOM Eternal & STG & 2 & Slime Rancher & SIM & 1 \\
Elefant Test FPS & STG & 4 & Stardew Valley & SIM & 1 \\
Fortnite & STG & 1 & Roblox: Debug (source label) & OTH & 1 \\
Helldivers 2 & STG & 1 & Eternal Return & OTH & 1 \\
Left 4 Dead 2 & STG & 2 & Goat Simulator & OTH & 1 \\
PUBG & STG & 1 & Goat Simulator 3 & OTH & 1 \\
Quake & STG & 2 & League of Legends & OTH & 2 \\
Rainbow Six & STG & 1 & Other / Unlabeled & OTH & 1 \\
Ready or Not & STG & 1 & Roblox: Be a Shark & OTH & 1 \\
Resident Evil 5 & STG & 1 & Roblox: Be a Snake & OTH & 1 \\
Resident Evil: Revelations 2 & STG & 1 & Roblox: Be a Tornado & OTH & 1 \\
Roblox: Hypershot & STG & 1 & Roblox: Blade Ball & OTH & 1 \\
Roblox: Murderers vs Sheriffs & STG & 2 & Roblox: Death Ball & OTH & 3 \\
Roblox: Rivals & STG & 2 & Roblox: Eat the World & OTH & 1 \\
VALORANT & STG & 1 & Roblox: Evade & OTH & 2 \\
Warhammer: Vermintide & STG & 1 & Roblox: Slap Battle & OTH & 2 \\
Warhammer: Vermintide 2 & STG & 1 & Rocket League & OTH & 1 \\
Cyberpunk 2077 & RPG & 1 & Super Bunny Man & OTH & 1 \\
GTA V & RPG & 2 & Vampire Survivors & OTH & 1 \\
\bottomrule
\end{tabular}}
\end{table}

\textbf{Validation sets.}
We use three fixed held-out sets to evaluate action loss. The Minecraft set consists of trajectories from the VPT test split. The in-corpus set is game-balanced and draws from D2E and OpenP2P recordings held out from the S4--S5 construction, together with a size-matched Minecraft subset from the VPT test split. The out-of-distribution (OOD) set contains only game titles absent from every S1--S5 mixture. All three sets use the same observation and action representation as training. Recordings and adjacent overlapping windows are kept separate across training and validation to prevent leakage.

\textbf{OOD source and construction.}
The OOD set is drawn from the processed Gaming500~\citep{markov2026gaming500} recording pool, which is not used to construct any S1--S5 training mixture. We first remove exact-title overlaps with VPT, D2E, and OpenP2P, as well as obvious aliases, games from the same retained franchise, and emulator labels that resolve to a training title. We then retain keyboard--mouse recordings only. A recording is rejected when at least 5\% of its foreground events belong to a browser, desktop, launcher, or other non-game application; when at least half of its mouse events exhibit cursor recentering; or when its keyboard--mouse activity is too sparse (fewer than 60 input events per minute, fewer than three keyboard or click events per minute, or more than 60\% idle windows). We additionally require at least 60\% of a title's clean recordings to satisfy the keyboard--mouse criteria. These filters remove gamepad sessions whose labels are all zero, cursor-locked sessions whose mouse deltas are uninformative, and recordings dominated by menus or unrelated applications.

The resulting set contains 28 games and 4,480 clips. We sample exactly 160 non-idle clips per game, so every title contributes $1/28=3.57\%$ of the set, and distribute those clips as evenly as possible over two to four independent recording sessions (92 sessions in total). The set is therefore balanced by game rather than by the amount of available source video duration. Using the rendered interaction space as the taxonomy, it comprises 21 3D games (3,360 clips; 75.0\%), three 2.5D games (480 clips; 10.7\%), and four 2D games (640 clips; 14.3\%). We count isometric scenes rendered with 3D geometry, such as \emph{Dota 2}, as 3D; 2.5D denotes side-view or planar interaction with mixed 2D/3D presentation. Under the same game-category taxonomy used for the training inventory, the OOD set contains six AVG, five STG, three RPG, four RAC, four SIM, and six OTH titles. Table~\ref{tab:ood_validation_games} gives the complete composition.

\begin{table}[t]
\centering
\caption{\textbf{Composition of the OOD validation set.} All clips are drawn from Gaming500 recordings and come from titles absent from S1--S5. ``Sessions'' denotes the number of independent source recordings. Each game contributes exactly 160 clips.}
\label{tab:ood_validation_games}
\scriptsize
\setlength{\tabcolsep}{5pt}
\adjustbox{max width=\linewidth}{%
\begin{tabular}{rlllrr}
\toprule
\textbf{\#} & \textbf{Game} & \textbf{Category} & \textbf{Space} & \textbf{Sessions} & \textbf{Clips} \\
\midrule
1  & Palworld                                  & RPG & 3D   & 4 & 160 \\
2  & Assassin's Creed Syndicate                & AVG & 3D   & 4 & 160 \\
3  & 007 First Light                           & STG & 3D   & 4 & 160 \\
4  & Age of Empires II: Definitive Edition     & SIM & 2.5D & 4 & 160 \\
5  & Forza Horizon 5                           & RAC & 3D   & 4 & 160 \\
6  & Asphalt 9: Legends                        & RAC & 3D   & 4 & 160 \\
7  & PRAGMATA                                  & AVG & 3D   & 4 & 160 \\
8  & Candy Crush Saga                          & OTH & 2D   & 4 & 160 \\
9  & Dota 2                                    & OTH & 3D   & 4 & 160 \\
10 & The Finals                                & STG & 3D   & 4 & 160 \\
11 & Farming Simulator 18                      & SIM & 3D   & 4 & 160 \\
12 & Wolfenstein: The Old Blood                & STG & 3D   & 4 & 160 \\
13 & Hogwarts Legacy                           & RPG & 3D   & 4 & 160 \\
14 & Just Cause 2                              & AVG & 3D   & 2 & 160 \\
15 & ARC Raiders                               & STG & 3D   & 2 & 160 \\
16 & 2XKO                                      & OTH & 2.5D & 4 & 160 \\
17 & Terraria                                  & AVG & 2D   & 3 & 160 \\
18 & Pok\'emon Z (fan game)                    & RPG & 2D   & 2 & 160 \\
19 & Township                                  & SIM & 2D   & 3 & 160 \\
20 & South of Midnight                         & AVG & 3D   & 4 & 160 \\
21 & Ori and the Will of the Wisps             & OTH & 2.5D & 2 & 160 \\
22 & Free Fire MAX                             & STG & 3D   & 4 & 160 \\
23 & Beach Buggy Racing                        & RAC & 3D   & 4 & 160 \\
24 & Gear.Club Unlimited 2                     & RAC & 3D   & 2 & 160 \\
25 & Pacify                                    & AVG & 3D   & 2 & 160 \\
26 & ONE PIECE Bounty Rush                     & OTH & 3D   & 2 & 160 \\
27 & Schedule I                                & SIM & 3D   & 2 & 160 \\
28 & Dead as Disco                             & OTH & 3D   & 2 & 160 \\
\midrule
\multicolumn{3}{r}{\textbf{Total}} & \textbf{21 3D / 3 2.5D / 4 2D} & \textbf{92} & \textbf{4,480} \\
\bottomrule
\end{tabular}}
\end{table}

\subsection{Sparse Retrospective Reasoning Annotations and Action Template}
\label{app:prompts}

Qwen3.5 natively places deliberation within a \texttt{<think>} span before producing its final output. We preserve this structure across all CPT configurations because the same behavior frequently occurs during online gameplay evaluation. Retaining the reasoning channel keeps the training response format aligned with the reason-then-act format used at evaluation.

\textbf{Annotation procedure.}
For selected gameplay windows, GPT-5.6-Luna receives the recorded visual context together with the action that the player subsequently executed. It is instructed to adopt the player's perspective and reconstruct a short retrospective explanation of the apparent observation, intent, and chosen action. The annotation describes the recorded behavior rather than proposing or relabeling an action, so the original keyboard--mouse supervision remains unchanged. We inject these rationales sparsely among ordinary gameplay turns and apply the same procedure to S1--S5, preventing reasoning density from becoming an unintended diversity variable. Table~\ref{tab:reasoning_density} reports the resulting density for each mixture. A 20-frame training sample carries 1.08--1.19 non-empty rationales on average, 74--78\% of samples contain at least one, and only 5.4--6.0\% of assistant turns include reasoning. Each rationale averages 64.5--65.4 tokens under the Qwen3.5 tokenizer (pooled median 65; 10th--90th percentile 56--74), or 70--78 reasoning tokens per sample. Because rationale count, coverage, and length differ only marginally across S1--S5, reasoning supervision does not act as an additional variable alongside data diversity.

\begin{table}[!htbp]
\centering
\small
\caption{\textbf{Sparse reasoning density in the S1--S5 training mixtures.} Counts are from a full scan of the training parquets. A rationale is a non-empty \texttt{<think>} span attached to an assistant turn. Token lengths are measured with the Qwen3.5 tokenizer on 13k--19k sampled rationales per mixture.}
\label{tab:reasoning_density}
\begin{tabular}{lrrrrrr}
\toprule
Mixture & Samples & \begin{tabular}[b]{@{}r@{}}Samples w/\\rationale\end{tabular} & \begin{tabular}[b]{@{}r@{}}Rationales\\/ sample\end{tabular} & \begin{tabular}[b]{@{}r@{}}Reasoning\\turns\end{tabular} & \begin{tabular}[b]{@{}r@{}}Tokens\\/ rationale\end{tabular} & \begin{tabular}[b]{@{}r@{}}Reasoning tokens\\/ sample\end{tabular} \\
\midrule
S1 & 1.50M & 74.0\% & 1.08 & 5.40\% & 65.0 & 70.2 \\
S2 & 1.50M & 75.8\% & 1.14 & 5.69\% & 64.5 & 73.4 \\
S3 & 1.50M & 76.7\% & 1.17 & 5.83\% & 65.3 & 76.1 \\
S4 & 1.50M & 77.6\% & 1.19 & 5.96\% & 65.4 & 78.0 \\
S5 & 1.50M & 77.6\% & 1.19 & 5.97\% & 64.6 & 77.1 \\
\bottomrule
\end{tabular}
\end{table}

\textbf{Training format.}
\begin{quote}
\texttt{User: <image>} \\
\texttt{Assistant: <think>} \\
\texttt{[Sparse reasoning, when available]} \\
\texttt{</think>} \\
\texttt{<|action\_start|> $\Delta x$ $\Delta y$ $\Delta z$ ; $k_1$ ; \dots\ ; $k_6$ <|action\_end|>}
\end{quote}

\begin{figure}[!htbp]
\centering
\includegraphics[width=\linewidth]{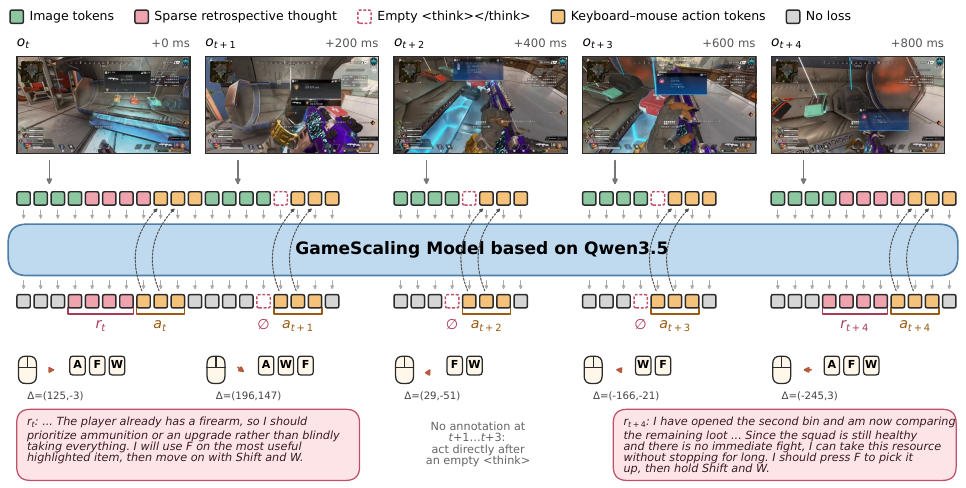}
\caption{\textbf{A training window with sparse reasoning.} Five consecutive 200\,ms turns from an Apex Legends recording in S4. Each turn begins with the current frame. Only $o_t$ and $o_{t+4}$ carry a GPT-5.6-Luna rationale; the other three turns emit an empty \texttt{<think>} span and act directly. Every action packet contains a mouse displacement $(\Delta x, \Delta y)$ and the keys held during six 33\,ms sub-steps; key icons show the distinct keys in each packet. Image tokens receive no loss, while thought and action tokens are trained autoregressively. Rationales are abridged excerpts of the stored annotations.}
\label{fig:reasoning_example}
\end{figure}

When no retrospective annotation is attached, the chat template supplies an empty \texttt{<think>} span, keeping the response grammar identical across examples. At evaluation, the model is free to generate a non-empty reasoning segment before the action packet. Figure~\ref{fig:reasoning_example} shows five consecutive turns from an S4 training window, two with rationales and three without.

\subsection{Full Accounting of the 120-Condition Factorial Grid}
\label{app:grid_accounting}

Table~\ref{tab:wsd_trunks} outlines the 20 primary Warmup--Stable--Decay (WSD) trunks~\citep{hu2024minicpm}. For every evaluated checkpoint, the reported gameplay-hour budget $T\in\{50,100,200,400,800,1600\}$ includes all three phases: warmup, stable training, and cosine decay. Thus, the 50-hour model has a total budget of 50 hours, including its decay stage. For each model--mixture pair, the shorter-budget models branch from earlier trunk checkpoints and complete their cosine decay when the cumulative budget reaches $T$. The 1,600-hour condition uses the trunk's own final decay endpoint. This reuse produces six completed checkpoints per model--mixture pair and 120 conditions in total.

\begin{table}[!htbp]
\centering
\small
\caption{\textbf{Factorial 120-condition experimental matrix.} Twenty primary WSD trunks, each producing five annealed branch checkpoints and one final trunk endpoint at total budgets $T \in \{50\mathrm{h}, 100\mathrm{h}, 200\mathrm{h}, 400\mathrm{h}, 800\mathrm{h}, 1600\mathrm{h}\}$. Each budget includes warmup, stable training, and decay.}
\label{tab:wsd_trunks}
\adjustbox{max width=\linewidth}{%
\begin{tabular}{lcccc}
\toprule
\textbf{Data Mixture ID} & \textbf{Composition Breakdown} & \textbf{Model Scales} & \textbf{Trunks} & \textbf{Evaluated Checkpoints} \\
\midrule
S1 (Pure Single-Source) & 100\% Minecraft & \{0.8B, 2B, 9B, 27B\} & 4 & 24 \\
S2 (Equal Mixture) & 51.4\% Minecraft & \{0.8B, 2B, 9B, 27B\} & 4 & 24 \\
S3 (Heterogeneous-Dominant) & 27.1\% Minecraft & \{0.8B, 2B, 9B, 27B\} & 4 & 24 \\
S4 (Broad Multi-Game) & 2.8\% Minecraft & \{0.8B, 2B, 9B, 27B\} & 4 & 24 \\
S5 (Minecraft-Excluded) & 0\% Minecraft & \{0.8B, 2B, 9B, 27B\} & 4 & 24 \\
\midrule
\textbf{Total Grid} & & & \textbf{20 Trunks} & \textbf{120 Conditions} \\
\bottomrule
\end{tabular}}
\end{table}

\subsection{Training Hyperparameters and Optimization Stability}
\label{app:hyperparameters}

All CPT runs use bf16 AdamW~\citep{loshchilov2019decoupled} ($\beta_1=0.9$, $\beta_2=0.95$, weight decay $0.01$) with gradient clipping at 1.0 and a fixed target batch budget of 2,359,296 tokens. The language-model peak learning rate is $1\times10^{-5}$ at every parameter scale. The vision encoder is unfrozen and uses a separate rate of $7\times10^{-6}$. The WSD schedule~\citep{hu2024minicpm} allocates 3\% of training to linear warmup, 87.3\% to the stable phase, and the final 9.7\% to cosine decay with a minimum learning-rate ratio of $0.1$. All three phases are included in the reported training budget, as defined in Appendix~\ref{app:grid_accounting}. Every condition uses $854\times480$ frames, a 20-frame gameplay context, a fixed 15\% vision--language token share, and seed 3407.

\section{Evaluation Details}
\label{app:evaluation}

\subsection{Benchmark Evaluation Environments}
\label{app:benchmarks}

\textbf{Level 1: Source-World Benchmarks (Minecraft and ViZDoom).}
MCU~\citep{zheng2025mcu} uses 30 long-horizon Minecraft tasks spanning navigation, survival, GUI interaction, combat, and crafting. ViZDoom~\citep{Wydmuch2019ViZdoom} evaluates first-person control on four skill-1 DOOM maps; the scoring protocol is given in Appendix~\ref{app:vizdoom_protocol}. Both environments are represented in the gameplay corpus, but at opposite densities: Minecraft spans 0--100\% of the gameplay budget across S1--S5, whereas footage of the classic DOOM titles on whose engine ViZDoom runs never exceeds one to two percent of any mixture; later DOOM releases in the corpus (e.g., DOOM Eternal) share the first-person shooting skill but differ in art style and rules. ViZDoom therefore probes a rare world and a frequent skill: DOOM itself is minimally exposed, but the first-person aiming and firing it tests is common to the games that dominate the heterogeneous pool, which is why the main text treats the pool, not DOOM, as its source.

\textbf{Level 2: Related-World Benchmark (OmniGameArena - OGA).}
Consists of 7 diverse photorealistic Unreal Engine 5 games. Scenarios cover third-person exploration, tactical combat, target shooting, and obstacle platforming, running on native UE5 execution environments.

\textbf{Level 3: Distant OOD Benchmark (GameWorld - GW).}
Comprises 170 tasks across 34 environments, mostly 2D and 2.5D, including classic arcade platformers, puzzle solvers (2048, Sokoban), and word-guessing games, whose visual styles, physics, and control logic depart from the training worlds.

\textbf{Action-interface adaptation.}
OGA natively consumes a packet similar to the \texttt{<|action\_start|>}~\ldots~\texttt{<|action\_end|>} format described in Section~\ref{sec:setup}; we set its mouse-displacement range and the number of keys per substep to match ours, and change nothing else. MCU, ViZDoom, and GameWorld do not. For each of these three benchmarks we add a thin adapter that parses the packet into the harness's native keyboard--mouse events: the six per-substep key sets are replayed as a 6$\times$33\,ms key timeline, and the mouse displacement is mapped to the environment's native camera or cursor control. The adapter changes only the action interface. Task definitions, observation resolution, episode length, step caps, success criteria, scoring, and every other evaluation parameter remain those of the original benchmark, and the same adapter is used for every model we evaluate in Table~\ref{tab:cross_benchmark_reference}, including the frontier models, which receive the packet grammar through the same prompt and are parsed by the same code.

\subsection{Evaluation Protocol}
\label{app:eval_protocol}

\textbf{Episodes and averaging.}
Every reported score is a mean over tasks of the mean over episodes. MCU runs each of its 30 tasks 5 times and report the mean and standard deviation over 5 runs in Table~\ref{tab:mcu_model_comparison}; OGA runs each of its 7 games 10 times; ViZDoom runs each of its 4 maps 10 times; GameWorld runs each of its 170 tasks once. The same counts apply to every model and every checkpoint in the grid, and to every model in Table~\ref{tab:cross_benchmark_reference}. The average score on the four benchmarks is reported in Table~\ref{tab:cross_benchmark_reference}.

\textbf{Provenance of the two comparison tables.}
The prior-model rows of Table~\ref{tab:mcu_model_comparison} are quoted from their source papers. Their backbones, action interfaces, training data, and harnesses are therefore not aligned with ours, and the table should be read as a reference to the published state of Minecraft agents rather than as a controlled comparison. Table~\ref{tab:cross_benchmark_reference} is different: we evaluated every model in it ourselves, under our action interface, with the adapters above, the same prompts, the same observation resolution and history, and the same episode counts. Only the model changes between rows.

\textbf{Episode budgets and termination.}
Each benchmark keeps its native budget. MCU allows a fixed 600 step budget in the time-paused setting and 3 minutes of game time per task in the real time setting. ViZDoom and GameWorld use a fixed step cap. OGA ends an episode when either its game-time limit or its step cap is reached, whichever comes first. On MCU, ViZDoom, and OGA an episode stops immediately when the task fails or the agent dies, and the episode is scored as it stands. On GameWorld, failure or death restarts the game, and the agent keeps playing until the step cap is reached; the score is the progress accumulated within the cap. Under real-time interaction, the difference is that the world advances while the model reasons.

\subsection{ViZDoom Scoring Protocol}
\label{app:vizdoom_protocol}

\textbf{Maps and per-map score.}
We use four maps at \texttt{doom\_skill} 1: \texttt{battle1} and \texttt{battle2} from Direct Future Prediction, and the standard \texttt{defend\_the\_center} and \texttt{defend\_the\_line} scenarios. The two families carry different native units. On the battle maps we record the frag count, following the convention of the work that introduced them. On the defend maps we record the cumulative engine reward, which awards $+1$ per kill and $-1$ on death. The battle WADs also expose a per-kill reward through ACS global 0, and we verified over 3{,}358 episodes that it equals the frag count on every one, so the two conventions coincide there and a paper reporting ``average episode reward'' on these maps reports the same quantity.

\textbf{Aggregate score.}
The reported ViZDoom number is the unweighted mean of the four per-map scores after each is divided by 100:
\[
S_{\text{ViZDoom}} \;=\; \frac{1}{4}\sum_{m \in \mathcal{M}} \frac{\bar{s}_m}{100},
\qquad
\mathcal{M} = \{\texttt{battle1}, \texttt{battle2}, \texttt{dtc}, \texttt{dtl}\},
\]
where $\bar{s}_m$ averages 10 episodes on map $m$.

\subsection{Serving and Deployment Configurations}
\label{app:deployment}

All GameScaling checkpoints are deployed in bf16 on an NVIDIA H200 cluster and evaluated through vLLM API endpoints. The context length is sized to the observation history, \texttt{limit-mm-per-prompt} equals the history length plus one, sampling uses temperature 1.0, and generation stops at \texttt{<|action\_end|>}. Under time-paused interaction, every MCU, ViZDoom, OGA, and GameWorld evaluation job reserves eight H200 GPUs and serves the model through vLLM. Under real-time interaction, where inference latency enters the score, the 0.8B and 2B models use tensor-parallel size 1 on one H200 per vLLM instance, while the 9B and 27B models use tensor-parallel size 4 on four H200 GPUs per instance. Table~\ref{tab:deployment} summarizes these protocol-dependent layouts.

\begin{table}[!htbp]
\centering
\small
\caption{\textbf{Serving layouts for the GameScaling models.} All time-paused benchmark jobs reserve eight H200 GPUs. Real-time vLLM instances use TP\,=\,1 for 0.8B/2B and TP\,=\,4 for 9B/27B so that the reported score includes the latency of the stated deployment.}
\label{tab:deployment}
\adjustbox{max width=\linewidth}{%
\begin{tabular}{lllll}
\toprule
\textbf{Protocol} & \textbf{Benchmarks} & \textbf{Model sizes} & \textbf{vLLM deployment} & \textbf{Timing semantics} \\
\midrule
Time-paused & MCU, ViZDoom, OGA, GameWorld & 0.8B, 2B, 9B, 27B & 8 H200 GPUs per job & Environment frozen during inference \\
Real-time &  MCU, ViZDoom, OGA, GameWorld  & 0.8B, 2B & TP\,=\,1, one H200 per instance & Environment advances during inference \\
Real-time &  MCU, ViZDoom, OGA, GameWorld  & 9B, 27B & TP\,=\,4, four H200 GPUs per instance & Environment advances during inference \\
\bottomrule
\end{tabular}}
\end{table}

\textbf{Time-paused interaction.}
Because the environment is frozen while the model reasons, decision latency does not enter the score and requests may be scheduled for throughput. Every time-paused run on MCU, ViZDoom, OGA, and GameWorld uses an eight-H200 job with vLLM serving. The environment backend remains benchmark-specific: Minecraft, ViZDoom, and Chromium execute outside the model server, while OGA uses UE5 instances on GPU. For OGA, each UE5 instance retains a dedicated GPU because packing multiple instances on one GPU starves real-time rendering during the 6$\times$33\,ms action chunk and can depress the score without producing an execution error.

\textbf{Real-time interaction.}
Under real-time interaction the measured latency of each decision is part of the score: the environment keeps advancing and the model keeps idle until the new one arrives. Actually, once model latency significantly exceeds 200 ms (such as 0.5s), both executing the previous action chunk and staying idle result in a severe performance degradation by more than $5\times$. Due to high latency, our 27B model also suffers an obvious performance drop. Each live worker therefore uses a dedicated vLLM endpoint rather than sharing a request queue. The 0.8B and 2B models are served on one H200 with TP\,=\,1; the 9B and 27B models are served on four H200 GPUs with TP\,=\,4. Splitting the 0.8B model across four GPUs raised its per-step latency from about 130\,ms to 178--190\,ms.

\textbf{Frontier models.}
Proprietary and hosted open-source models are evaluated through their provider APIs with the same adapter and prompt. All six models run at the default reasoning effort documented by their providers, which we do not override: \texttt{medium} for GPT-5.6-Sol, GPT-5.6-Luna, and Gemini-3.8-Flash; adaptive thinking at \texttt{high} effort for Claude-Opus-5.0; \texttt{high} for DeepSeek-V4-Flash-Vision; and \texttt{max} for GLM-5.3-Flash, whose thinking cannot be disabled. The output cap is 2,048 tokens on MCU, ViZDoom and OGA and 4,096 on GameWorld, the same for every model and both protocols. Models are run strictly one at a time with 5 concurrent workers under time-paused interaction and 2 under real-time interaction; their real-time scores therefore include provider-side latency, which is the latency a deployed agent would experience.

\section{Supplementary Experiments}
\label{app:supplementary}

\subsection{Frontier Progress by Release Date}
\label{app:frontier_trends}

\begin{figure}[!htbp]
\centering
\includegraphics[width=\linewidth]{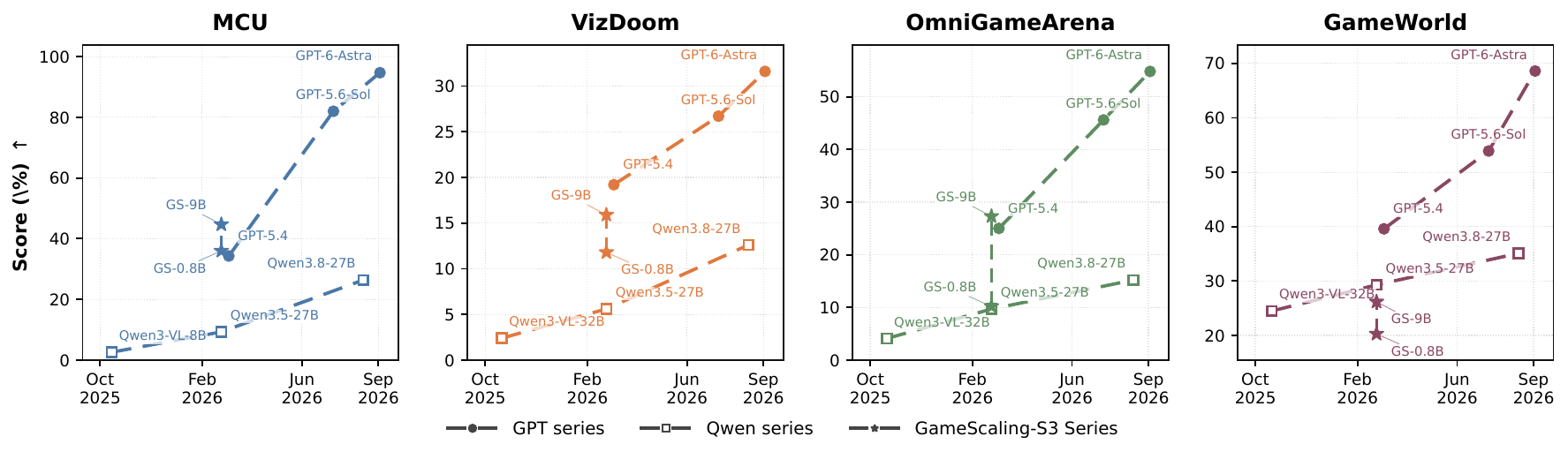}
\caption{\textbf{Frontier progress on the four benchmarks against release date.} Circles: GPT series; squares: Qwen series; stars: GameScaling S3 models at 1,600\,h, all under time-paused interaction.}
\label{fig:frontier_trends}
\end{figure}
\textbf{Progress in general models can bring larger gains than game-specific CPT.}
Figure~\ref{fig:frontier_trends} shows substantial improvements in gameplay across successive GPT and Qwen models under time-paused evaluation. Over the release range shown, gains in frontier-model performance can exceed those from our game-specific CPT on a fixed backbone. This difference is particularly clear on GameWorld, where CPT provides limited gains while newer frontier models improve markedly. These results suggest that general model capability is a major contributor to gameplay performance. Game-specific CPT develops useful skills, but improving the backbone itself remains an important route to stronger game agents.
\subsection{Loss Scaling Against Model Size}
\label{app:loss_model_axis}

\begin{figure*}[!htbp]
\centering
\includegraphics[width=\textwidth]{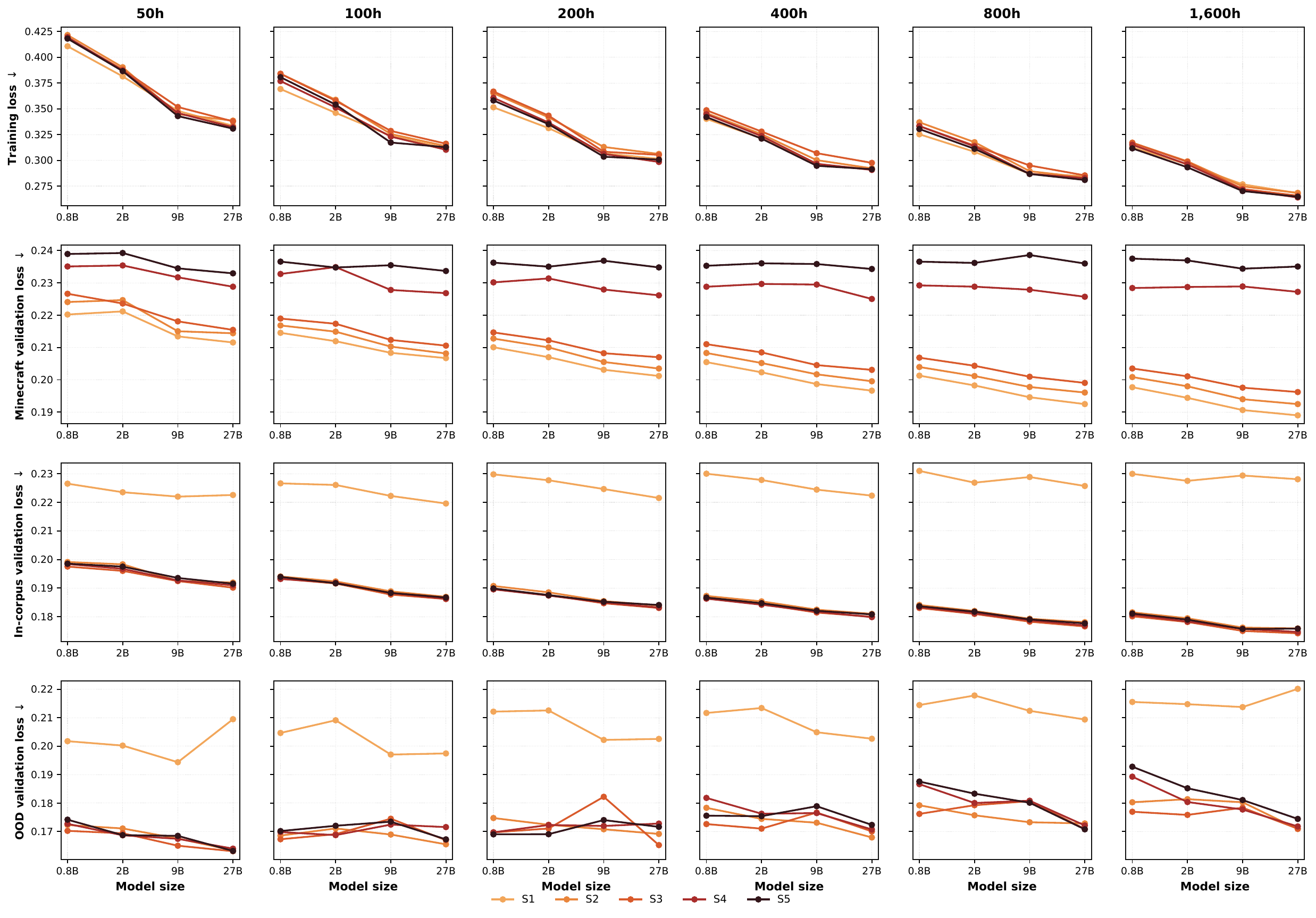}
\caption{\textbf{Loss scaling against model size.} Rows: training loss at the checkpoint, and held-out validation loss on Minecraft, on the game-balanced in-corpus set, and on the out-of-distribution set; columns: 50, 100, 200, 400, 800, and 1,600 gameplay hours; colors: mixtures S1--S5.}
\label{fig:loss_grid_model_axis}
\end{figure*}

Figure~\ref{fig:loss_grid_model_axis} replots the losses of Figure~\ref{fig:loss_scaling_grid} with model size on the horizontal axis and one column per duration. Across the six budgets, larger models generally have lower Minecraft and mixed-training in-corpus losses. OOD loss behaves differently: mixed training limits its later increase in larger models, while single-game training can weaken their advantage.

\subsection{Benchmark Scaling Against Gameplay Hours}
\label{app:hours_axis}

\begin{figure*}[!htbp]
\centering
\includegraphics[width=\textwidth]{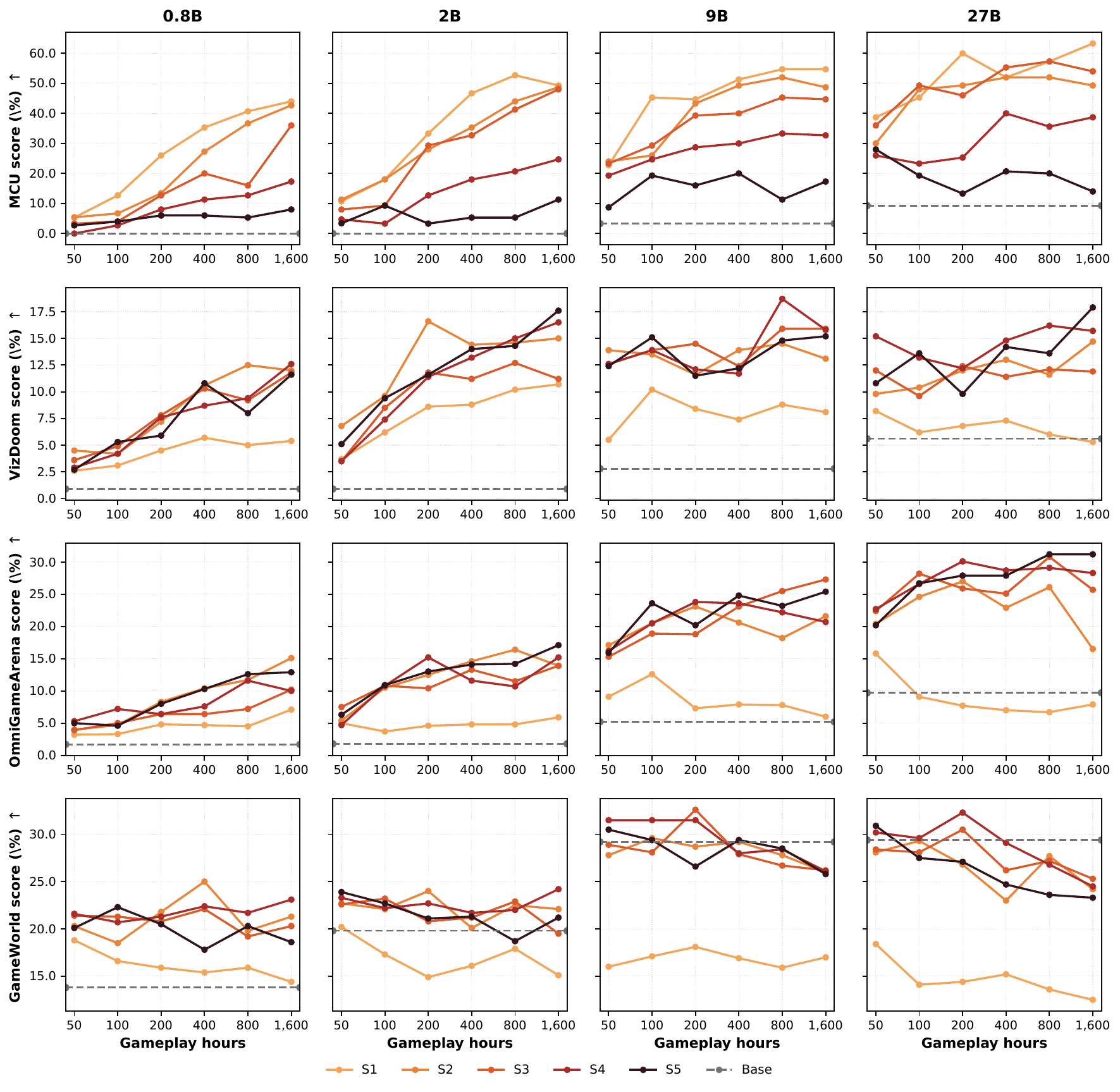}
\caption{\textbf{Benchmark scaling against gameplay hours under time-paused interaction.} Rows: MCU, ViZDoom, OGA, and GameWorld (all in \%); columns: 0.8B, 2B, 9B, and 27B backbones; points span 50 to 1,600 gameplay hours; colors: mixtures S1--S5; dashed gray: the untrained Qwen3.5 backbone.}
\label{fig:benchmark_grid_hours}
\end{figure*}

Figure~\ref{fig:benchmark_grid_hours} replots the 100 scores of Figure~\ref{fig:benchmark_grid} with gameplay hours on the horizontal axis and one column per model size, so the duration dynamics of Section~\ref{sec:data_scaling} can be read directly: the monotone gains of 0.8B and 2B under every mixture, the flattening of 9B and 27B after 100--400\,h, the decline of S1 at 9B and 27B on OGA and ViZDoom, and the decline of every 27B mixture on GameWorld.

\subsection{Test-Time Scaling on MCU}
\label{app:test_time_scaling}

\begin{figure}[!htbp]
\centering
\includegraphics[width=0.7\linewidth]{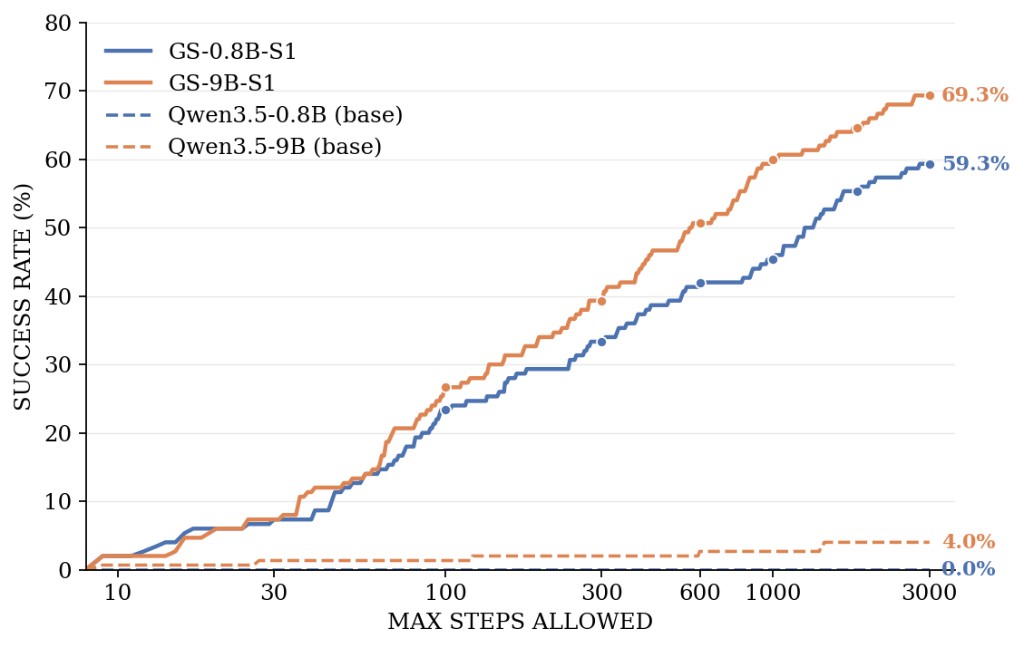}
\caption{\textbf{Test-time scaling on MCU.} Cumulative success rate of GameScaling-0.8B-S1 and GameScaling-9B-S1 (1,600\,h, solid) and their untrained Qwen3.5 backbones (dashed) on the 30 MCU tasks as a function of the maximum steps allowed, on a logarithmic axis. The standard protocol uses 600 steps.}
\label{fig:mcu_test_time_scaling}
\end{figure}

The standard MCU protocol caps every rollout at 600 steps, and all MCU numbers in the paper use this cap. Figure~\ref{fig:mcu_test_time_scaling} asks how much of the remaining failure is a step budget rather than a competence limit. We rerun GameScaling-0.8B-S1 and GameScaling-9B-S1 at 1,600\,h, together with their untrained Qwen3.5 backbones, with the cap raised to 3,000 steps and record, for each threshold, the fraction of the 150 rollouts that have succeeded within it. Success rises roughly log-linearly with the allowed steps for both models: from about 42\% and 51\% at the standard 600 steps to 59.3\% and 69.3\% at 3,000, with no sign of a plateau.  The two curves are nearly parallel, so the gain from more interaction steps is about the same for a 0.8B and a 9B model, and the size gap of about 10 points is preserved at every budget. The backbones do not scale this way. Qwen3.5-0.8B completes no task at any budget, and Qwen3.5-9B rises only from about 2\% at 600 steps to 4.0\% at 3,000. More steps help only a model that can already act in the world: CPT raises both the level and the test-time slope, and within the range we tested a GameScaling model given more time keeps converting it into completed tasks, independent of its size.

\subsection{Inference Latency and Output Length}
\label{app:latency_tokens}

The real-time columns of Table~\ref{tab:cross_benchmark_reference} compare API-served frontier models with GameScaling models on one or four H200 GPUs, and this setup does not favor GameScaling on hardware. The providers serve models of hundreds of billions to trillions of parameters on production clusters with far more compute than the single H200 behind GameScaling-0.8B and 2B or the four behind 9B and 27B. Even so, Table~\ref{tab:latency_tokens} shows frontier models taking 4.0--23.7\,s per step, against 0.13--0.69\,s for GameScaling; GameScaling-0.8B and 2B decide within the 200\,ms action chunk. The reason is output length: at their default reasoning efforts (Appendix~\ref{app:deployment}), counting hidden reasoning, visible \texttt{<think>} content, and the action, frontier models generate 123--1,027 tokens per step, 3 to 29 times the 35--39 of GameScaling, which learns sparse reasoning by continued pre-training on gameplay trajectories annotated with sparse rationales. Frontier models are slow in real time because they are large and reason at length before every action. Even the fastest of them takes about 20 times the 200\,ms action chunk per step. GameScaling acts within a second on one to four GPUs, which makes it the practical choice when both competence and latency count.

\begin{table}[!htbp]
\centering
\small
\caption{\textbf{Per-step inference latency and output length} of the Table~\ref{tab:cross_benchmark_reference} models, averaged over MCU, ViZDoom, OGA and GameWorld. Latency is the round trip of the model request, excluding screenshot capture and client-side image encoding; frontier models are called through the provider API, and GameScaling (S3, 1,600\,h) runs under real-time interaction on the deployment of Table~\ref{tab:deployment}. Output length is the \texttt{completion\_tokens} reported by the server, covering hidden reasoning, visible \texttt{<think>} content, and the action.}
\label{tab:latency_tokens}
\begin{tabular}{lrrr}
\toprule
\textbf{Model} & \textbf{Params} & \textbf{Latency (s/step)} & \textbf{Output tokens/step} \\
\midrule
Gemini-3.8-Flash & $\sim$T & 7.75 & 444 \\
GPT-5.6-Sol & $\sim$T & 5.57 & 123 \\
GPT-5.6-Luna & $\sim$T & 5.84 & 144 \\
Claude-Opus-5.0 & $\sim$T & 8.59 & 183 \\
GLM-5.3-Flash & 320B & 23.66 & 1,027 \\
DeepSeek-V4-Flash-Vision & 305B & 3.97 & 659 \\
\midrule
GameScaling-0.8B & 0.8B & 0.13 & 35 \\
GameScaling-2B & 2B & 0.16 & 35 \\
GameScaling-9B & 9B & 0.31 & 39 \\
GameScaling-27B & 27B & 0.69 & 36 \\
\bottomrule
\end{tabular}
\end{table}

\subsection{More Experiments on Scaling Law}
\label{app:scaling_law_experiments}

This section details the fits behind Section~\ref{sec:scaling_law}. We first describe the fitting procedure and check which losses decrease with training at all. We then derive Eq.~(\ref{eq:three_factor}) from the Chinchilla law, compare it with a categorical-mixture model, and report hold-out predictions and bootstrap uncertainty.

\textbf{Fitting procedure.}
Following Chinchilla~\citep{hoffmann2022compute}, we minimize the Huber loss of the log residual $\log\hat L-\log L$ with $\delta=10^{-3}$, so that errors are relative and single outliers do not dominate. The optimizer is L-BFGS-B, started from 192 initializations for Eq.~(\ref{eq:three_factor}) and 324 for the two-factor law, and the best optimum is kept. $E$, $A$, $B$ and $C$ are optimized in log space; the remaining parameters are bounded by $\alpha,\beta\in[0,3]$, $\eta\in[-3,3]$, $\gamma\in[0,5]$ and $\epsilon\in[10^{-4},1]$. $N$ is measured in billions of parameters and $T$ in gameplay hours. Unless a subset is stated, validation-loss laws are fitted to all 120 conditions of Figure~\ref{fig:loss_scaling_grid}, namely four sizes, five mixtures and six budgets from 50 to 1,600\,h. Each 50\,h validation point comes from its own annealing branch, like every other budget. The training loss is the 100-step EMA of the WSD trunk and is fitted at the five stable-phase budgets from 100 to 1,600\,h, giving 100 conditions. At 50\,h the trunk has just left its 3\% warmup, and its loss still falls faster than any power law in $T$.

Besides RMSE and $R^2$ over all conditions, we report the within-mixture $R^2_{\text{w}}$, which is the share of variance explained after the five mixture means are removed. Between-mixture differences account for 79\%, 88\% and 84\% of the total variance of the Minecraft, in-corpus and OOD losses. Any model with a mixture-dependent level therefore already reaches a high $R^2$, and $R^2_{\text{w}}$ measures what the $N$ and $T$ terms add on top of it. Two further points matter when reading the exponents. First, the objective is the Huber loss rather than $R^2$, so a parameter value that maximizes $R^2$ along a profile need not equal the estimate. Second, since $\rho$ takes five values, we draw only two qualitative conclusions from the $\rho$-dependent part of Eq.~(\ref{eq:three_factor}), namely that a world with $\rho=0$ has a finite floor and that $\eta>0$, and we do not interpret $\gamma$ or $\epsilon$ on their own.

\textbf{Which losses admit a scaling law.}
A power law in $T$ presupposes that loss falls with hours, so before fitting we check this on the raw curves, one per model size and mixture, or 20 per loss. On the Minecraft loss, 19 of the 20 curves end lower at 1,600\,h than at 50\,h, and all curves without a clear decrease belong to S5, which contains no Minecraft. On the in-corpus loss, the four rising curves are exactly those of S1, which contains no pool games. A world's loss thus decreases with hours whenever the world is in the corpus, and the exceptions are the mixtures that lack it. This is the pattern that the share $\rho$ and the floor term of Eq.~(\ref{eq:three_factor}) have to describe.

The OOD loss behaves differently in every cell. CPT lowers it from 0.24--0.31 for the backbone to about 0.17, but this gain is already realized at 50\,h, and further hours raise the loss again. All 20 curves end $0.007$--$0.019$ higher at 1,600\,h than at 50\,h, the rank correlation between hours and loss is positive in all 20 with a median of $0.91$, the loss rises over 76 of the 100 doublings, and the per-curve slope ranges from $+0.001$ to $+0.004$ per doubling. The OOD loss therefore violates the premise of a data scaling law before any law is fitted. We do not model its dynamics, because a form that can follow a rising curve, such as one with an added $D\log T$ term, would describe drift rather than scaling. Instead, we fit Eq.~(\ref{eq:three_factor}) to the OOD loss only as a negative control, to test whether a nine-parameter law manufactures a data term out of mixture differences alone. As shown below, it does not: $\beta$ goes to zero, as the raw curves imply it should.

\textbf{Fitted parameters.}
Table~\ref{tab:scaling_law_params} lists the point estimates behind the last column of Figure~\ref{fig:loss_scaling_grid}; their uncertainty is analysed at the end of this section. The reading that survives this analysis is structural: the floor term $C/(\rho+\epsilon)^{\gamma}$ produces the gap between mixtures with and without a world. On the Minecraft loss, going from $\rho=1$ to $\rho=0$ raises the floor term by $0.08$ and removes a data term worth $0.05$ at 100\,h. The net increase of $0.03$ matches the observed gap of $0.032$ between S5 and S1. On the in-corpus loss the corresponding numbers are $0.08$, $0.04$ and $0.04$, against an observed gap of $0.040$ between S1 and S2. Within mixtures, where the floor term is constant and only $N$ and $T$ act, the law still explains 91\% of the variance of the Minecraft loss and 96\% of the in-corpus loss, compared with 8\% of the OOD loss. The individual exponents are less certain. $\alpha$ is of order $0.2$ for both validation losses but has wide intervals, $\beta$ trades off against the offset $E$ along a ridge, and on the OOD loss $\beta$ sits on its lower bound.

\begin{table}[!htbp]
\centering
\small
\caption{\textbf{Parameters of the laws in Figure~\ref{fig:loss_scaling_grid}, column 5.} The validation losses follow Eq.~(\ref{eq:three_factor}) with the stated $\rho$ and are fitted on 120 conditions. The training loss follows $L=E_S+A/N^{\alpha}+B/T^{\beta}$, with $E_S$ between $0.005$ and $0.011$ across S1--S5, and is fitted on the 100 stable-phase conditions. $R^2_{\text{w}}$ is the within-mixture coefficient of determination.}
\label{tab:scaling_law_params}
\setlength{\tabcolsep}{4.5pt}
\adjustbox{max width=\linewidth}{%
\begin{tabular}{llrrrrrrrrrrrr}
\toprule
\textbf{Loss} & $\rho$ & $E$ & $A$ & $\alpha$ & $B$ & $\beta$ & $\eta$ & $C$ & $\gamma$ & $\epsilon$ & $R^2$ & $R^2_{\text{w}}$ & RMSE \\
\midrule
Training   & --    & $E_S$ & 0.069 & 0.39 & 0.400 & 0.07 & --   & --     & --   & --    & 0.980 & 0.980 & 0.0041 \\
Minecraft  & $r$   & 0.047 & 0.019 & 0.15 & 0.106 & 0.16 & 0.24 & 0.0987 & 0.13 & 0.011 & 0.982 & 0.912 & 0.0020 \\
In-corpus  & $1-r$ & 0.136 & 0.014 & 0.17 & 0.078 & 0.14 & 0.10 & 0.0031 & 1.00 & 0.039 & 0.995 & 0.959 & 0.0012 \\
OOD        & $1-r$ & 0.112 & 0.036 & 0.03 & 0.006 & 0.00 & 3.00 & 0.0213 & 0.26 & 0.014 & 0.855 & 0.079 & 0.0058 \\
\bottomrule
\end{tabular}}
\end{table}

\textbf{From Chinchilla to Eq.~(\ref{eq:three_factor}).}
The Chinchilla law $L=E+A/N^{\alpha}+B/D^{\beta}$ assumes that a single distribution is both trained on and evaluated, so that $D$ counts data from the scored distribution, and that loss falls monotonically with $D$. Our grid trains on mixtures and scores four distributions, which breaks the first assumption, and the OOD loss breaks the second. Like the Chinchilla and D-CPT laws, Eq.~(\ref{eq:three_factor}) is an empirical description rather than a law derived from a model of learning, and we make no claim beyond the grid. Its components are nevertheless not free design choices. The share $\rho$ is neither fitted nor tuned but fixed by how the mixtures are built (Appendix~\ref{app:dataset_construction}). The functional form is the published D-CPT law with its domain ratio replaced by $\rho$. Each term is forced by an observation that a simpler form cannot accommodate. We reached Eq.~(\ref{eq:three_factor}) by relaxing the Chinchilla assumptions one at a time on the Minecraft loss, each step guided by the residuals of the previous one (Table~\ref{tab:scaling_law_baselines}). The form is then tested against a categorical model of the same size and on held-out blocks, including held-out mixtures.
\begin{enumerate}
\setlength{\itemsep}{1pt}
\item \emph{Total hours, $D=T$.} On S1--S3 the fit reaches only $R^2=0.925$. The three mixtures share every value of $T$, yet their losses differ by up to $0.007$, which the law cannot separate. Total hours are therefore not the data variable of an evaluated world.
\item \emph{Minecraft hours, $D=rT$.} $R^2$ rises to $0.959$, but the residuals are ordered by mixture, from $+0.0016$ for S1 to $-0.0012$ for S3. At equal Minecraft hours, the mixture with more pool hours has the lower Minecraft loss, so hours on other worlds also lower a world's loss, at a lower rate.
\item \emph{Two data terms.} Adding a shared term in total hours, $L=E+A/N^{\alpha}+B/(rT)^{\beta}+C/T^{\gamma}$, removes the ordering and reaches $R^2=0.989$; the fitted $\gamma\approx0.5$ means that the shared term saturates early. This form is undefined at $r=0$, however, whereas S4 and S5 reach finite Minecraft losses of $0.229$ and $0.236$, far below the backbone. A world absent from training thus has a floor rather than an infinite loss, and the share must enter the law in a form that admits $\rho=0$.
\item \emph{Eq.~(\ref{eq:three_factor}).} The D-CPT form~\citep{dcpt2024}, with $D$ replaced by $T$ and the domain ratio by $\rho$, covers all five mixtures with one parameter set and reaches $R^2=0.982$. Its RMSE is at most $0.0015$ on S1--S3 and $0.003$ on S4--S5, where only two mixtures anchor the floor. The shares are the measured Minecraft trajectory shares of 100, 51.4, 27.1, 2.8 and 0\% from Appendix~\ref{app:dataset_construction}, not the nominal VPT fractions.
\item \emph{What the fitted form says.} $\eta$ is positive, $0.24$ for Minecraft and $0.10$ for the pool, so the data term does not reduce to effective hours $\rho T$. Forcing $\eta=-\beta$, which gives $B/(\rho T)^{\beta}$, lowers $R^2$ to $0.825$ on the Minecraft loss and $0.945$ on the in-corpus loss, and to $0.923$ and $0.953$ when the $\rho=0$ mixture is excluded. Constraining $\eta\le0$ on the $\rho>0$ conditions alone drives $\eta$ to the bound and lowers the Minecraft $R^2$ from $0.979$ to $0.934$. The sign of $\eta$ follows from the shape of the curves rather than from how the shares are labelled. Mixtures with small $\rho$ are nearly flat in $T$ while $\rho=1$ decays, so the amplitude of the $T$ term must grow with $\rho$. At 0.8B, for example, the S5 Minecraft loss moves only from $0.239$ to $0.237$ over 50--1,600\,h, while the S1 loss falls from $0.220$ to $0.198$. This is step~2 seen from inside the final law: the floor term, not the data term, carries the mixture dependence.
\item \emph{Second evaluation world.} With $\rho=1-r$ the same form describes the in-corpus loss, while the ranking of the restricted baselines inverts. $D=T$ fits S2--S5 well, with $R^2=0.991$, because the four diverse mixtures have almost identical in-corpus losses. $D=(1-r)T$ is worse, at $0.955$, because Minecraft hours also lower the loss on pool games. Only Eq.~(\ref{eq:three_factor}) admits S1, where $\rho=0$. S1 remains the largest residual, with RMSE $0.0024$ against at most $0.0007$ for the other mixtures, because its in-corpus loss rises with hours, which no monotone law can reproduce.
\item \emph{Training and OOD loss.} The training loss is scored on each mixture's own stream, so $\rho$ is undefined and the law reduces to Chinchilla with a per-mixture offset $E_S$. On the OOD loss monotonicity already fails in the raw curves, so Eq.~(\ref{eq:three_factor}) is applied there only as a negative control. Replacing it with a form that fits a rising loss would no longer yield a scaling law.
\end{enumerate}
Two alternatives were set aside. A weighted sum of two Chinchilla laws, one per validation slice with $D=rT$ and $D=(1-r)T$, is easier to interpret but diverges at $r=0$ and $r=1$, so it covers only S2--S4. Treating the number of worlds as a variable cannot be tested on this grid, because S2--S5 share one fixed pool.

\textbf{The Minecraft slice of the in-corpus set.}
The in-corpus set contains 30 pool games and Minecraft with 160 clips each, so Minecraft makes up $\pi=1/31=3.2\%$ of it and $\rho=1-r$ is an approximation. To measure its effect, we write the measured loss as $L_{\text{in}}=\pi L_{\text{mc}}+(1-\pi)L_{\text{pool}}$ and remove the Minecraft slice using the measured Minecraft loss. Relative to $L_{\text{in}}$, this shifts $L_{\text{pool}}$ by $+0.0007$ for S1 and by $-0.0007$ to $-0.0017$ for S2--S5. The shift is largest for S4 and S5, whose Minecraft loss is highest, so the slice slightly narrows the gap between S1 and the diverse mixtures. Fitting Eq.~(\ref{eq:three_factor}) with $\rho=1-r$ to $L_{\text{pool}}$ gives $R^2=0.995$ and the same per-mixture RMSEs. Combining the two laws as $\pi\hat L_{\text{mc}}+(1-\pi)\hat L_{\text{pool}}$ predicts $L_{\text{in}}$ with $R^2=0.995$ and RMSE $0.0013$. Replacing $\rho$ by the validation-weighted share $\pi r+(1-\pi)(1-r)$ fits worse, with $R^2=0.989$, so the Minecraft slice enters additively rather than through the share.

\begin{table}[!htbp]
\centering
\small
\caption{\textbf{Restricted scaling-law baselines against the three-factor law.} Each baseline drops the share $\rho$ and keeps only $N$ and one data variable $D$ (the two-factor Chinchilla law); ``two-phase'' adds a second data term in total hours. $n$ is the number of fitted conditions.}
\label{tab:scaling_law_baselines}
\setlength{\tabcolsep}{5pt}
\begin{tabular}{lllrrr}
\toprule
\textbf{Loss} & \textbf{Law} & \textbf{Conditions} & $n$ & $R^2$ & RMSE \\
\midrule
Minecraft & Chinchilla, $D=T$ (total hours)        & S1--S3 & 72  & 0.925 & 0.0024 \\
Minecraft & Chinchilla, $D=rT$ (Minecraft hours)   & S1--S3 & 72  & 0.959 & 0.0018 \\
Minecraft & Two-phase, $D=rT$ and $T$              & S1--S3 & 72  & 0.989 & 0.0009 \\
Minecraft & Eq.~(\ref{eq:three_factor}), $\rho=r$ & S1--S5 & 120 & 0.982 & 0.0020 \\
\midrule
In-corpus & Chinchilla, $D=T$                      & S2--S5 & 96  & 0.991 & 0.0006 \\
In-corpus & Chinchilla, $D=(1-r)T$ (pool hours)    & S2--S5 & 96  & 0.955 & 0.0014 \\
In-corpus & Eq.~(\ref{eq:three_factor}), $\rho=1-r$ & S1--S5 & 120 & 0.995 & 0.0012 \\
\midrule
OOD       & Chinchilla, $D=T$                      & S2--S5 & 96  & 0.117 & 0.0055 \\
OOD       & Eq.~(\ref{eq:three_factor}), $\rho=1-r$ & S1--S5 & 120 & 0.855 & 0.0058 \\
\bottomrule
\end{tabular}
\end{table}

\textbf{Eq.~(\ref{eq:three_factor}) against a categorical-mixture model.}
With only five levels on the mixture axis, the natural alternative to a parametric $\rho$ is to treat the mixture as a factor. Table~\ref{tab:scaling_law_categorical} compares Eq.~(\ref{eq:three_factor}) with the \emph{categorical offset} model $L=E_S+A/N^{\alpha}+B/T^{\beta}$, which gives every mixture its own offset and has the same nine parameters as Eq.~(\ref{eq:three_factor}). Both are fitted with the same procedure on the same 120 conditions. We compare them by Gaussian AIC and BIC on the loss residuals and by blocked cross-validation, holding out one block of the grid at a time: the 27B models, the 1,600\,h checkpoints, the 50\,h checkpoints, or one mixture.

The comparison supports two conclusions. First, the mixture effect is not a mere offset. The categorical offset model has about twice the RMSE of Eq.~(\ref{eq:three_factor}) on both losses, $0.0044$ against $0.0020$ on Minecraft and $0.0035$ against $0.0012$ in-corpus. Its $R^2_{\text{w}}$ drops to $0.56$ and $0.64$, and BIC prefers Eq.~(\ref{eq:three_factor}) by $190$ and $260$. What the offset misses is that the slope in $T$ depends on the share. The Minecraft loss falls by $0.0045$ per doubling of hours under S1, by $0.0013$ under S4 and not at all under S5, and the in-corpus loss falls by $0.0035$ per doubling under S2--S5 but rises under S1. Eq.~(\ref{eq:three_factor}) captures this interaction through $\eta$ and the floor term.

Second, only Eq.~(\ref{eq:three_factor}) can predict a mixture it has not seen, because $\rho$ enters as a covariate rather than a label. It predicts a held-out S2 or S3 with RMSE $0.0020$--$0.0021$ on the Minecraft loss and $0.0008$--$0.0016$ on the in-corpus loss, whereas the categorical model has no parameter for such a fold. With the same number of parameters, Eq.~(\ref{eq:three_factor}) therefore both fits better and generalizes across mixtures.

\begin{table}[!htbp]
\centering
\small
\caption{\textbf{Eq.~(\ref{eq:three_factor}) against a categorical-mixture model of the same size.} Both models are fitted on the 120 conditions with the Huber-log objective, and $k$ is the number of parameters. AIC and BIC use the Gaussian likelihood of the loss residuals, where lower is better. CV columns give the RMSE on the held-out block after refitting on the rest; ``--'' marks folds on which the model has no parameter for the unseen mixture.}
\label{tab:scaling_law_categorical}
\setlength{\tabcolsep}{4pt}
\adjustbox{max width=\linewidth}{%
\begin{tabular}{llrrrrrrrrrrrr}
\toprule
\textbf{Loss} & \textbf{Model} & $k$ & $R^2$ & $R^2_{\text{w}}$ & RMSE & AIC & BIC & CV 27B & CV 1,600\,h & CV 50\,h & CV S2 & CV S3 & CV S4 \\
\midrule
Minecraft & Eq.~(\ref{eq:three_factor}), $\rho=r$             & 9  & 0.982 & 0.912 & 0.0020 & $-1135$ & $-1110$ & 0.0026 & 0.0027 & 0.0025 & 0.0020 & 0.0021 & 0.0057 \\
Minecraft & Categorical offset $E_S$                          & 9  & 0.909 & 0.562 & 0.0044 & $-942$  & $-917$  & 0.0049 & 0.0071 & 0.0070 & -- & -- & -- \\
\midrule
In-corpus & Eq.~(\ref{eq:three_factor}), $\rho=1-r$           & 9  & 0.995 & 0.959 & 0.0012 & $-1257$ & $-1232$ & 0.0016 & 0.0018 & 0.0022 & 0.0016 & 0.0008 & 0.0005 \\
In-corpus & Categorical offset $E_S$                          & 9  & 0.958 & 0.636 & 0.0035 & $-996$  & $-971$  & 0.0037 & 0.0051 & 0.0059 & -- & -- & -- \\
\bottomrule
\end{tabular}}
\end{table}

\textbf{Negative control on the OOD loss.}
Since the OOD loss rises with hours in every cell, a law with a decreasing data term should find no such term, and both laws behave accordingly. The two-factor law fitted to S2--S5 reaches $R^2=0.117$ with $B\to0$ and degenerates to a function of size only. Eq.~(\ref{eq:three_factor}) raises the nominal $R^2$ to $0.855$, but only through the floor term, which absorbs the $0.035$ gap between S1 and the diverse mixtures. Within mixtures it explains 8\% of the variance, $\beta$ is driven to its lower bound in 97\% of the bootstrap replicates of Table~\ref{tab:scaling_law_uncertainty}, and the predicted loss is flat in $T$. The residuals therefore carry the rise seen in the raw curves. Averaged over sizes, the observed loss increases by $0.002$--$0.003$ per doubling under S2--S5, from $0.0016$ at 27B to $0.0031$ at 0.8B, and by $0.003$--$0.004$ under S1. $\beta\approx0$ is thus exactly the outcome the control is designed to detect, not a failure to fit: the flexibility of Eq.~(\ref{eq:three_factor}) does not create a data-scaling term where the data contain none. Within 50--1,600\,h, CPT hours do not scale the loss on worlds absent from the corpus.

\textbf{Hold-out predictions.}
Table~\ref{tab:scaling_law_holdout} refits Eq.~(\ref{eq:three_factor}) on the grid with one block of conditions held out and predicts that block, covering all three axes of the grid. Along the size axis, a fit on 0.8B--9B predicts the 27B models with RMSE $0.0026$ on the Minecraft loss and $0.0016$ on the in-corpus loss. Along the data axis the law extrapolates in both directions. A fit on 50--800\,h predicts the 1,600\,h checkpoints with RMSE $0.0027$ and $0.0018$, and a fit on 100--1,600\,h predicts the 50\,h checkpoints with RMSE $0.0025$ and $0.0022$. These errors are $1.3$--$1.8$ times the in-sample RMSE, and the mean residuals of these size and data blocks stay below $0.0014$, so the extrapolated predictions carry no systematic bias. Along the diversity axis, the law interpolates to a held-out S2 or S3 with RMSE $0.0020$--$0.0021$ on the Minecraft loss and $0.0008$--$0.0016$ on the in-corpus loss. Leaving out S4 is harder for the Minecraft loss, with RMSE $0.0057$, because S5 is then the only mixture near $\rho=0$; the in-corpus loss of S4 is still predicted with RMSE $0.0005$. Eq.~(\ref{eq:three_factor}) therefore interpolates between the observed shares and extrapolates about one doubling in model size and in either direction of the data budget.

\begin{table}[!htbp]
\centering
\small
\caption{\textbf{Hold-out predictions of Eq.~(\ref{eq:three_factor}).} Each row refits the law on all conditions outside the held-out block and evaluates it on the block. The in-sample RMSE on all 120 conditions is $0.0020$ for the Minecraft loss and $0.0012$ for the in-corpus loss. Held-out S4 Minecraft losses span a narrow range, so $R^2_{\text{test}}$ is not informative there and is omitted.}
\label{tab:scaling_law_holdout}
\setlength{\tabcolsep}{5pt}
\adjustbox{max width=\linewidth}{%
\begin{tabular}{lllrrrrr}
\toprule
\textbf{Axis} & \textbf{Fit $\rightarrow$ predict} & \textbf{Loss} & $n_{\text{fit}}$ & $n_{\text{test}}$ & $R^2_{\text{test}}$ & RMSE & Mean residual \\
\midrule
Size      & 0.8B--9B $\rightarrow$ 27B          & Minecraft & 90  & 30 & 0.971 & 0.0026 & $+0.0013$ \\
          &                                     & In-corpus & 90  & 30 & 0.991 & 0.0016 & $+0.0008$ \\
\midrule
Data      & 50--800\,h $\rightarrow$ 1,600\,h   & Minecraft & 100 & 20 & 0.978 & 0.0027 & $+0.0003$ \\
          &                                     & In-corpus & 100 & 20 & 0.992 & 0.0018 & $+0.0004$ \\
          & 100--1,600\,h $\rightarrow$ 50\,h   & Minecraft & 100 & 20 & 0.916 & 0.0025 & $+0.0007$ \\
          &                                     & In-corpus & 100 & 20 & 0.967 & 0.0022 & $+0.0009$ \\
\midrule
Diversity & S1,S3--S5 $\rightarrow$ S2          & Minecraft & 96  & 24 & 0.945 & 0.0020 & $+0.0013$ \\
          &                                     & In-corpus & 96  & 24 & 0.939 & 0.0016 & $+0.0015$ \\
          & S1,S2,S4,S5 $\rightarrow$ S3        & Minecraft & 96  & 24 & 0.929 & 0.0021 & $-0.0008$ \\
          &                                     & In-corpus & 96  & 24 & 0.986 & 0.0008 & $-0.0007$ \\
          & S1--S3,S5 $\rightarrow$ S4          & Minecraft & 96  & 24 & --    & 0.0057 & $-0.0021$ \\
          &                                     & In-corpus & 96  & 24 & 0.993 & 0.0005 & $-0.0002$ \\
\bottomrule
\end{tabular}}
\end{table}

\textbf{Uncertainty of the fits.}
The grid has a single training seed, so the resampling below measures how much each fit depends on which conditions enter it, given the measured losses. It does not measure the seed-to-seed variance of the losses themselves. We use two schemes with $B=300$ replicates each and refit every replicate from the full-data estimate plus eight perturbed starts. The \emph{stratified case bootstrap} resamples conditions with replacement within each mixture, so every replicate keeps all five mixtures. Without stratification a replicate can lose most of S5, the only mixture at $\rho=0$, and $\gamma$ and $\epsilon$ then diverge for reasons unrelated to the model. The \emph{residual bootstrap} keeps every $(N,T,\rho)$ and resamples the fitted residuals. It preserves the factorial design but assumes exchangeable residuals, which the systematic S1 in-corpus residual violates. Table~\ref{tab:scaling_law_uncertainty} reports 95\% percentile intervals from the case scheme. The residual scheme gives intervals of similar width for every parameter of the training, Minecraft and in-corpus laws. For hold-out errors, only the fitted conditions are resampled, and the held-out conditions stay fixed.

\begin{table}[!htbp]
\centering
\small
\caption{\textbf{Uncertainty of the three-factor laws.} Each entry is the point estimate with a 95\% percentile interval over $B=300$ stratified case-bootstrap replicates. ``RMSE on grid'' evaluates each replicate law on the original conditions and thus measures how much the predicted surface moves. ``Stable $\beta$'' is the range of fixed $\beta$ for which refitting the other parameters keeps $R^2$ within $0.001$ of the unconstrained fit, together with the resulting range of $E$. Hold-out rows refit each law on the stated conditions and report RMSE and mean residual on the fixed held-out conditions; their point estimates are the fits of Tables~\ref{tab:scaling_law_categorical} and~\ref{tab:scaling_law_holdout}.}
\label{tab:scaling_law_uncertainty}
\setlength{\tabcolsep}{4pt}
\adjustbox{max width=\linewidth}{%
\begin{tabular}{llll}
\toprule
& \textbf{Training} & \textbf{Minecraft} & \textbf{In-corpus} \\
\midrule
$\alpha$              & 0.39 [0.31, 0.53] & 0.15 [0.02, 0.30] & 0.17 [0.06, 0.30] \\
$\beta$               & 0.070 [0.064, 0.078] & 0.16 [0.12, 0.27] & 0.14 [0.08, 0.20] \\
$\eta$                & -- & 0.24 [0.18, 0.35] & 0.10 [0.03, 0.18] \\
$\gamma$              & -- & 0.13 [0.10, 0.26] & 1.00 [0.86, 1.82] \\
$E$                   & $E_{S1}$ 0.005 [0.001, 0.009] & 0.047 [0.013, 0.114] & 0.136 [0.095, 0.149] \\
$R^2$                 & 0.980 [0.973, 0.986] & 0.982 [0.975, 0.988] & 0.995 [0.993, 0.997] \\
RMSE on grid          & 0.0041 [0.0042, 0.0049] & 0.0020 [0.0019, 0.0024] & 0.0012 [0.0012, 0.0013] \\
Stable $\beta$ (profile) & [0.07, 0.12], $E_{S1}\in[0.003, 0.11]$ & [0.15, 0.50], $E\in[0.04, 0.15]$ & [0.03, 0.30], $E\in[0.01, 0.15]$ \\
\midrule
Hold-out $\le$9B $\rightarrow$ 27B: RMSE & 0.0092 [0.0059, 0.0129] & 0.0026 [0.0019, 0.0035] & 0.0016 [0.0014, 0.0019] \\
\hspace{1.2em} mean residual              & $+0.0078$ [$+0.0019$, $+0.0117$] & $+0.0013$ [$+0.0004$, $+0.0022$] & $+0.0008$ [$+0.0003$, $+0.0012$] \\
Hold-out $\le$800\,h $\rightarrow$ 1,600\,h: RMSE & 0.0092 [0.0065, 0.0141] & 0.0027 [0.0023, 0.0039] & 0.0018 [0.0017, 0.0023] \\
\hspace{1.2em} mean residual              & $-0.0081$ [$-0.0128$, $-0.0048$] & $+0.0003$ [$-0.0007$, $+0.0013$] & $+0.0004$ [$-0.0001$, $+0.0011$] \\
Hold-out 100--1,600\,h $\rightarrow$ 50\,h: RMSE & -- & 0.0025 [0.0023, 0.0034] & 0.0022 [0.0019, 0.0024] \\
\hspace{1.2em} mean residual              & -- & $+0.0007$ [$-0.0002$, $+0.0014$] & $+0.0009$ [$+0.0006$, $+0.0011$] \\
\bottomrule
\end{tabular}}
\end{table}

\begin{figure*}[!htbp]
\centering
\includegraphics[width=\textwidth]{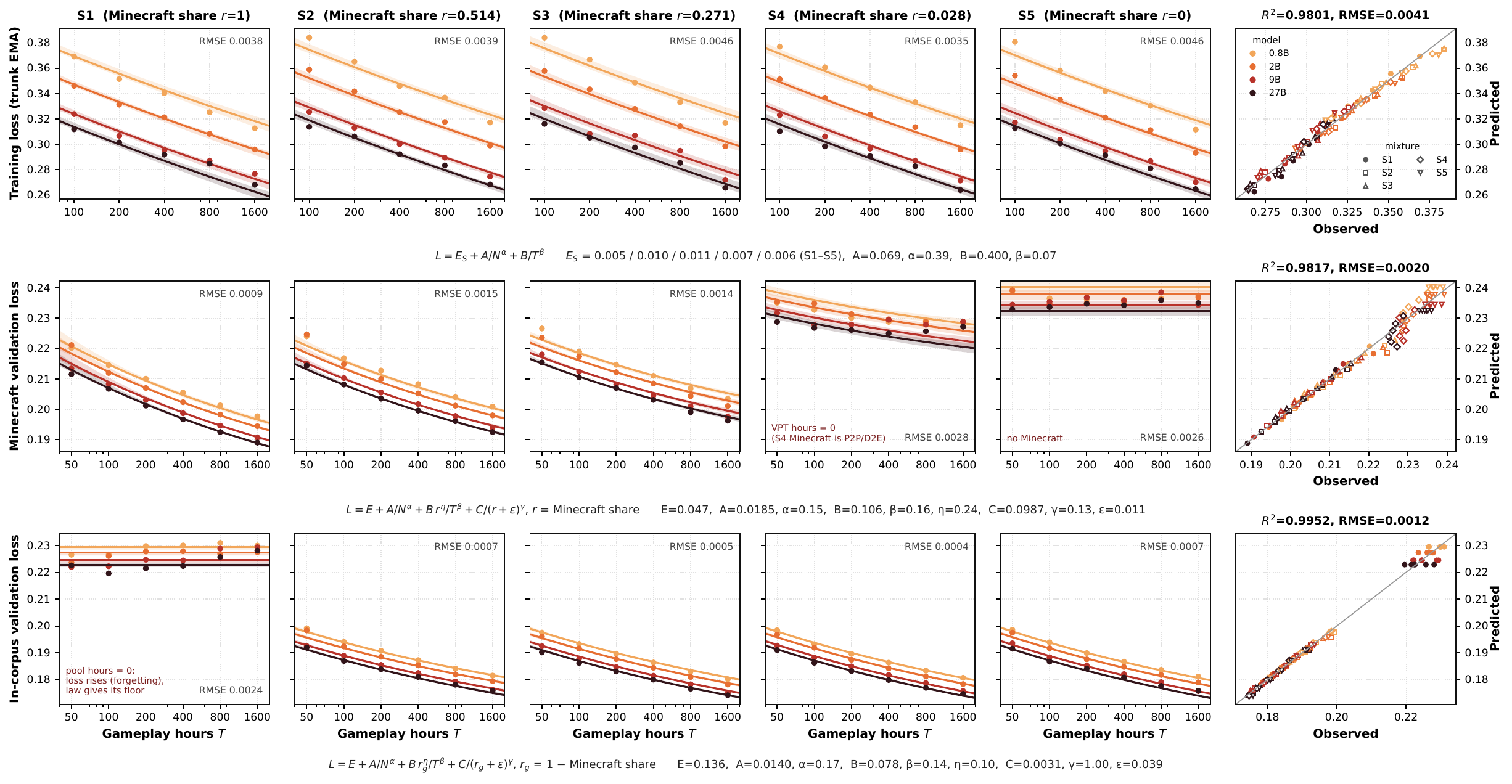}
\caption{\textbf{Three-factor laws with prediction bands.} Rows are the training loss, the Minecraft validation loss and the in-corpus validation loss; columns are the mixtures S1--S5; colors are model sizes. The horizontal axis is gameplay hours, from 100\,h for the training loss and from 50\,h for the validation losses. Points are measurements, lines are the single fitted law of each row evaluated at that panel's $(N,r)$, and shaded bands are the 5--95\% range of the law over $B=300$ stratified case-bootstrap replicates. The last column is predicted against observed loss; per-panel RMSE is printed.}
\label{fig:scaling_law_grid_3x5}
\end{figure*}

Four conclusions follow. First, the predicted surfaces are stable even where individual parameters are not. Evaluated on the original grid, the replicate laws change the RMSE by at most $0.0004$ for the Minecraft law and $0.0001$ for the in-corpus law, and the 5--95\% bands in Figure~\ref{fig:scaling_law_grid_3x5} are narrower than the markers over most of the grid.

Second, four model sizes determine $\alpha$ only loosely. Its Minecraft interval spans $0.02$ to $0.30$ because $E$, $A$ and $\alpha$ compensate each other, so we only state that capacity enters with an exponent of order $0.2$ and do not compare $\alpha$ across losses.

Third, $\beta$ and $E$ lie on a ridge. Fixing $\beta$ anywhere between $0.15$ and $0.5$ for Minecraft, or between $0.03$ and $0.3$ for the in-corpus loss, and refitting the rest keeps $R^2$ within $0.001$ of the unconstrained fit, while $E$ moves by $0.11$--$0.14$. Along the Minecraft ridge $R^2$ even peaks at $\beta\approx0.3$, above the Huber estimate of $0.16$, as expected for a flat ridge under two different criteria. The asymptotic floor is therefore not a quantity this grid can determine, and we do not report it. For the training law the ridge is one-sided, with $\beta$ between $0.07$ and $0.12$. Below $0.07$ the offsets $E_S$ reach their lower bound and $R^2$ drops steeply, so the estimate sits at the constrained end of the ridge, and we likewise do not interpret $E_S$ as an asymptote.

Fourth, the hold-out errors of the two validation laws exceed their in-sample errors by a factor of $1.3$--$1.8$, and their mean residuals lie within one in-sample RMSE of zero in both directions along the data axis. One doubling of size or hours is therefore a fair extrapolation range. The training law extrapolates worse, with RMSE $0.009$ in both directions, and the signs of its residuals show why. The 1,600\,h point is the end of the WSD decay, which lowers the trunk loss faster than any power law in $T$, and the 27B loss is higher than an $\alpha$ fitted on 0.8B--9B predicts. On the OOD loss, finally, $\beta$ sits at its lower bound in 97\% of case-bootstrap replicates while $\eta$ and $\gamma$ remain unconstrained, so the absence of a data term is a property of the data rather than of one optimizer run. Under the residual scheme the fraction drops to 51\%, as expected, because resampling the residuals destroys exactly the monotone trend in $T$ that they carry.

\section{Comparison with Related Scaling Studies}
\label{app:cpt_scale_comparison}

GameScaling studies CPT across 120 conditions, combining four model sizes, six data budgets, and five world mixtures. We compare its training scale with language-model scaling studies, and its model range, data coverage, and experimental scope with GUI and robotics studies cited in Section~\ref{sec:related_work}.

\subsection{Language-Model Scaling Studies}

D-CPT Law~\citep{dcpt2024} and CMR Scaling Law~\citep{cmr2024} study how data mixtures affect domain-adaptive CPT, while \emph{When Data Is Scarce}~\citep{sparse_scaling2026} studies sparse models trained from scratch with repeated data. As shown in Table~\ref{tab:cpt_scale_comparison}, GameScaling's reported $2.2\times10^{22}$ FLOPs are about $4.7\times$ the estimated CMR budget and of the same order as D-CPT and the dense-equivalent estimate for \emph{When Data Is Scarce}. Our controlled grid extends to 27B parameters and connects training scale with both held-out loss and performance in interactive worlds.

\begin{table}[!htbp]
\centering
\small
\caption{\textbf{Training scale of language-model studies and GameScaling.} Budgets cover the experiments listed, excluding original backbone pre-training for CPT. Related-work budgets are estimates; GameScaling uses the reported totals in Section~\ref{sec:setup}.}
\label{tab:cpt_scale_comparison}
\setlength{\tabcolsep}{5pt}
\renewcommand{\arraystretch}{1.18}
\begin{tabularx}{\linewidth}{@{}>{\raggedright\arraybackslash}p{0.25\linewidth}>{\raggedright\arraybackslash}Xcc@{}}
\toprule
\textbf{Study} & \textbf{Experiments covered} & \textbf{Tokens} & \textbf{FLOPs} \\
\midrule
D-CPT Law\newline\citep{dcpt2024} & 0.5--4B; 6 domains;\newline 9 mixtures (main grid) & $4.2\times10^{12}$\textsuperscript{$\dagger$} & $5.3\times10^{22}$\textsuperscript{$\dagger$} \\
\addlinespace[4pt]
CMR Scaling Law\newline\citep{cmr2024} & 0.46--3.1B; 29 Finance and Academic Papers runs & $5.8\times10^{11}$\textsuperscript{$\dagger$} & $4.7\times10^{21}$\textsuperscript{$\dagger$} \\
\addlinespace[4pt]
When Data Is Scarce\newline\citep{sparse_scaling2026} & Up to 1.92B dense-equivalent; 500 fitting runs & $\sim3.3\times10^{12}$\textsuperscript{$\ddagger$} & $\sim1.2\times10^{22}$\textsuperscript{$\ddagger$} \\
\midrule
\textbf{GameScaling} & 0.8--27B; 120 conditions;\newline 6 budgets, 5 mixtures & $3.5\times10^{11}$ & $2.2\times10^{22}$ \\
\bottomrule
\end{tabularx}
\par\smallskip
\raggedright\footnotesize
$\dagger$ Estimated from published training settings; D-CPT counts the nominal grid before possible reuse of general-only runs. $\ddagger$ Rough estimate from \href{https://github.com/boqian333/sparse-dc-scaling/blob/main/fitting/fitting.ipynb}{500 fitting records} with balanced model, data, and repetition settings; dense-equivalent FLOPs (ideal sparse: $\sim3.7\times10^{21}$). Larger-model extrapolation tests are excluded. Estimates indicate scale, not an exact cost ranking.
\end{table}

\subsection{GUI and Robotics Scaling Studies}
\label{app:embodied_scale_comparison}

Table~\ref{tab:embodied_scale_comparison} compares model sizes, data coverage, experimental variables, and training resources. GUI studies cover models up to 32B and large collections of interface examples or trajectories. The robotics studies listed here focus on data diversity, robot coverage, or training design at a fixed model size. GameScaling combines a broad model range with controlled changes in data volume and world mixture. Its approximately 45,000 H200 GPU-hours support a common protocol for measuring both scaling within training worlds and transfer to unseen related and distant worlds.

\begin{table}[!htbp]
\centering
\small
\caption{\textbf{Scope of GUI, robotics, and virtual-world scaling studies.} GameScaling cost covers the full grid; other costs are per run. $\dagger$: time estimate at unchanged recipe and throughput. Data use the original units; GameScaling hours include warmup, stable training, and decay.}
\label{tab:embodied_scale_comparison}
\setlength{\tabcolsep}{3.5pt}
\renewcommand{\arraystretch}{1.18}
\begin{tabularx}{\linewidth}{@{}>{\raggedright\arraybackslash}p{0.20\linewidth}>{\raggedright\arraybackslash}p{0.11\linewidth}>{\raggedright\arraybackslash}p{0.22\linewidth}>{\raggedright\arraybackslash}p{0.17\linewidth}>{\raggedright\arraybackslash}X@{}}
\toprule
\textbf{Study} & \textbf{Models} & \textbf{Data scale} & \textbf{Main variables} & \textbf{Training resources} \\
\midrule
ScaleCUA\newline\citep{scalecua2026} & 3B, 7B, 32B & 17.1M grounding;\newline 471K understanding;\newline 19K trajectories & Model size;\newline data mixture & 128 A100 (3B/7B);\newline 128 H200 (32B) \\
\addlinespace[5pt]
GUIMid\newline\citep{guimid2025} & 7B & Up to 300K examples;\newline ${\sim}$55K GUI samples & Task type;\newline mixture; volume & 8 GPUs;\newline 1 epoch \\
\addlinespace[5pt]
OpenThoughts-Agent\newline\citep{datarecipes2026} & 8B, 32B & 316--100K trajectories & Data quality;\newline volume;\newline $>100$ ablations & GH200 GPU-h:\newline 160 (10K),\newline ${\sim}$1,600\textsuperscript{$\dagger$} (100K);\newline 8B SFT \\
\midrule
Imitation data scaling\newline\citep{ilscaling2025} & $>396$M & $>40$K demonstrations;\newline 4 tasks & Environments;\newline objects; demos & 8 A800;\newline ${\sim}$64\textsuperscript{$\dagger$}--600 GPU-h \\
\addlinespace[5pt]
Rethinking VLA\newline\citep{rethinkingvla2026} & 2B +\newline 0.7B & 182.5M transitions & Action format;\newline schedule;\newline regularization & 8 A800;\newline 40K+200K updates \\
\addlinespace[5pt]
Cross-Embodiment Offline RL\newline\citep{crossembodiment2026} & URMA & 16M transitions/variant;\newline 16 robots & Data quality;\newline robot grouping & 100K updates;\newline 5 seeds \\
\midrule
\textbf{GameScaling} & 0.8--27B & 50--1,600 hours;\newline 5 world mixtures & Model size;\newline data volume;\newline world mixture & 45K H200 GPU-h;\newline 120 conditions \\
\bottomrule
\end{tabularx}
\par\smallskip
\raggedright\footnotesize
VLA sizes list the VLM and action expert separately. GPU-hours across different device types are not directly comparable.
\end{table}

\end{document}